\documentclass[letterpaper]{article} % DO NOT CHANGE THIS
\usepackage{aaai2026}  % DO NOT CHANGE THIS
\usepackage{times}  % DO NOT CHANGE THIS
\usepackage{helvet}  % DO NOT CHANGE THIS
\usepackage{courier}  % DO NOT CHANGE THIS
\usepackage[hyphens]{url}  % DO NOT CHANGE THIS
\usepackage{graphicx} % DO NOT CHANGE THIS
\usepackage{natbib}  % DO NOT CHANGE THIS AND DO NOT ADD ANY OPTIONS TO IT
\usepackage{caption} % DO NOT CHANGE THIS AND DO NOT ADD ANY OPTIONS TO IT
\usepackage{algorithm}
\usepackage{algorithmic}
\usepackage{booktabs}
\usepackage{multirow}
\usepackage{amssymb}
\usepackage{amsmath}
\usepackage{pdfpages}
\usepackage{newfloat}
\usepackage{listings}
\DeclareCaptionStyle{ruled}{labelfont=normalfont,labelsep=colon,strut=off} % DO NOT CHANGE THIS
\floatstyle{ruled}
\newfloat{listing}{tb}{lst}{}
\floatname{listing}{Listing}
\title{Token Painter: Training-Free Text-Guided \\ Image Inpainting via Mask Autoregressive Models}
\author{
    Longtao Jiang\textsuperscript{\rm 1},
    Jie Huang\textsuperscript{\rm 1},
    Mingfei Han\textsuperscript{\rm 2},
    Lei Chen\textsuperscript{\rm 1}, \\
    Yongqiang Yu\textsuperscript{\rm 2},
    Feng Zhao\textsuperscript{\rm 1},
    Xiaojun Chang\textsuperscript{\rm 1},
    Zhihui Li\textsuperscript{\rm 1\dag}
}
\affiliations{
    \textsuperscript{\rm 1}University of Science and Technology of China \ 
    \textsuperscript{\rm 2}Department of Computer Vision, MBZUAI \\
    \{taotao707,hj0117\}@mail.ustc.edu.cn, \
    {mingfei.han,yongqiang.yu}@mbzuai.ac.ae \ \\
    chenlei.hfut@gmail, \
    \{fzhao956,xjchang,lizhihuics\}@ustc.edu.cn \\
}

\usepackage{bibentry}
\begin{document}

\twocolumn[{%
\renewcommand\twocolumn[1][]{#1}%
\maketitle
\begin{center}
  \centering
  \vspace{-1.5em}
  \captionsetup{type=figure}
   \includegraphics[width=1\linewidth]{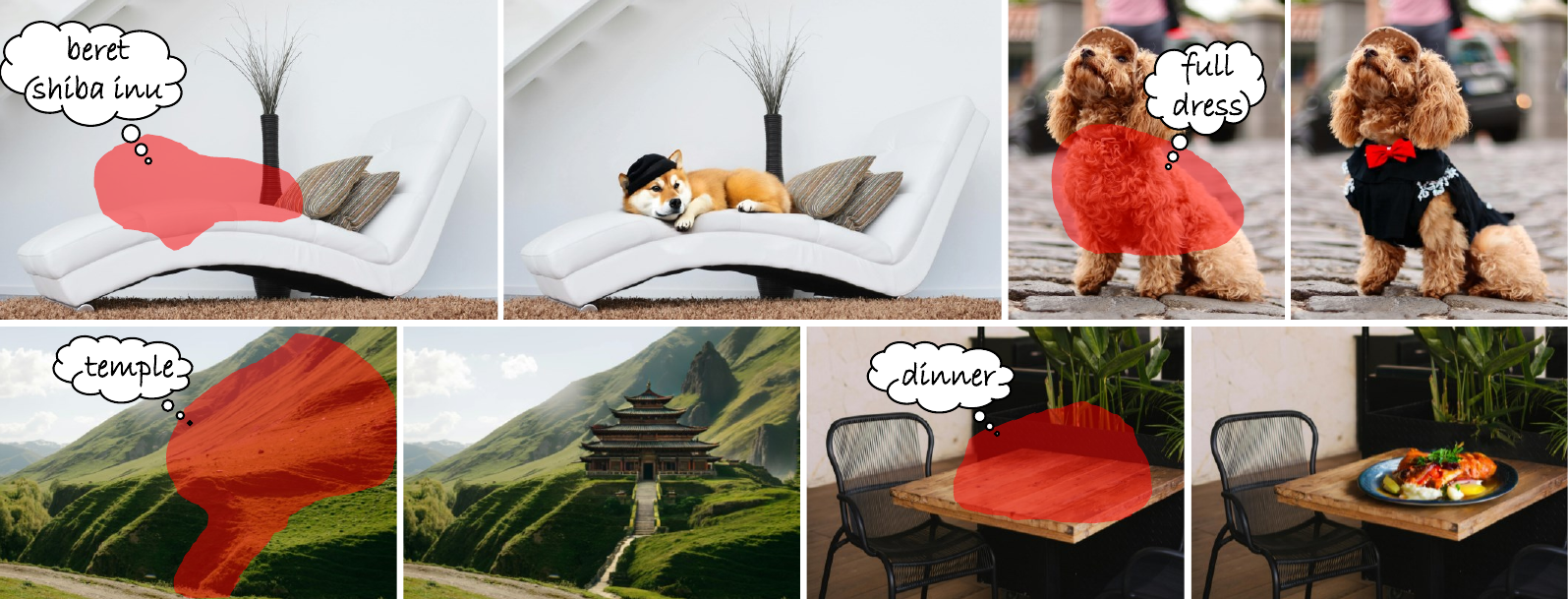}
   \vspace{-1.5em}
   \caption{Text-guided image inpainting results generated by our method Token Painter. It faithfully follows the prompt details to fill the masked region while seamlessly connecting with the image context.}
   \label{fig:teasor}
\end{center}
}]

\def\thefootnote{\dag}\footnotetext{\noindent Zhihui Li is the corresponding author.}\def\thefootnote{\arabic{footnote}}

%定义一些概念性的东西，不要出现过多的专业名词，比如什么机制，只说大体的内容概念，不要说方法。
% Text-guided image inpainting是旨在根据给定的prompt来无缝的重绘目标区域，同时保持背景部分的内容不变。随着最近基于diffusion model的文生图技术的发展，stable diffusion inpainting（SDI）及其改进工作逐渐在此任务中占据主导地位。然而由于SDI在训练时采取了random mask的训练策略，其根据text生成的物体质量天然受限。同时，diffusion model在潜在空间从纯噪声中生成整张图片的特性，造成了背景部分一致性的损害及不必要计算资源的浪费。为了解决这些问题，本文基于Mask Autoregressive（MAR）模型进行了text-guided image inpainting任务的研究。特别的，MAR模型可以在潜在空间中灵活的生成指定部分的token而不改变其他token，因此其天然适合image inpainting任务，更进一步，由于局部物体生成的意图与其在训练时的目标一致，生成的物体质量天然不受限。我们首先在MAR模型上进行了两种简单的inpainting方法的实现，然而结果分别显示了其忽视prompt和与背景的不和谐，通过深入研究注意力图和潜在token的特性，我们发现了text和image两种模态token的作用机制。利用这些发现，我们提出了Token Painter，一种基于MAR的Training-free的Text-guided image inpainting方法，其在MAR的encoder和decoder中进行两阶段调整。首先，它通过Dual-Stream Encoder Information Fusion （DEIF）模型在encoder阶段对两种模态token的信息进行了融合，使在其主导下，inpaiting部分生成的内容不仅可以准确的follow prompt同时还可以与背景部分保持和谐。然后，我们在decoder阶段使用了Adaptive Decoder Attention Score Enhancing（ADAE）模块，通过在生成过程中动态调整交叉注意力和自注意力的分数，进一步提升了对prompt细节的遵循以及物体生成质量。广泛的实验表明，Training-free的Token Painter几乎在所有的指标上都超过了现有的SOTA方法且展现出了杰出的视觉效果，即使这些方法是基于专门在image inpainting数据集上进行训练的模型。我们将在未来开源所有代码，模型及Benchmark。

\begin{abstract}
Text-guided image inpainting aims to inpaint masked image regions based on a textual prompt while preserving the background. Although diffusion-based methods have become dominant, their property of modeling the entire image in latent space makes it challenging for the results to align well with prompt details and maintain a consistent background.
To address these issues, we explore Mask AutoRegressive (MAR) models for this task. MAR naturally supports image inpainting by generating latent tokens corresponding to mask regions, enabling better local controllability without altering the background. However, directly applying MAR to this task makes the inpainting content either ignore the prompts or be disharmonious with the background context. Through analysis of the attention maps from the inpainting images, we identify the impact of background tokens on text tokens during the MAR generation, and leverage this to design \textbf{Token Painter}, a training-free text-guided image inpainting method based on MAR.
Our approach introduces two key components: (1) Dual-Stream Encoder Information Fusion (DEIF), which fuses the semantic and context information from text and background in frequency domain to produce novel guidance tokens, allowing MAR to generate text-faithful inpainting content while keeping harmonious with background context. (2) Adaptive Decoder Attention Score Enhancing (ADAE), which adaptively enhances attention scores on guidance tokens and inpainting tokens to further enhance the alignment of prompt details and the content visual quality. Extensive experiments demonstrate that our training-free method outperforms prior state-of-the-art methods across almost all metrics. Codes: \url{https://github.com/longtaojiang/Token-Painter}.
\end{abstract}


% 第一段介绍image inpainting任务背景（对应你的方法符合任务特性，暗示只需要修复mask部分，保存背景部分不变），然后说随着t2i模型的发展，在其中text inpainting占这重要作用，应用等。DF在上面的效果逐渐起了主导作用。第二段说到DF在其中的潜在的不足，然后一些工作怎么缓解的，但是还是不能彻底缓解（委婉）（与任务特性不符合，全部重新生成，背景部分改变）。第三段最近AR在生成领域受到了越来越多的关注，讲一下AR在图像生成的发展，生成token的过程，及几种变体，然后开始切入MAR，最近的一些进展，讲一下他的机制和DF相呼应（灵活性，可以任意生成对应位置的token，效果，适合inpainting任务），因此MAR是天然适合于inpainting任务的，因此我们的工作目标就是探究MAR在inpainting的应用，此处插入intro图a。第四段就是探究过程，首先试一下两个极端，但是在实践过程中发现了两个分支的问题，我们将其归咎于text inpainting和generation本身的gap，提出假设，接下来验证了注意力图和统计量，揭示了存在的问题。第五段然后就开始说我们这边方法的内容了，对应方法部分的二个点，这两个模块的关系要写出来，对应第四段发现的问题，然后就开始介绍效果，此处插入intro图b。最后就是几点贡献：第一点我们首先将MAR用在inpainting，并且深入研究了MAR在inpainting的机制，第二点就是融合text和bg方面，第三点attention方面模块，第四点就是效果，可以加粗。
% Introduction放个图，一是MAR和DF的inpainting概念对比，二是质量和效果的不同方法的对比，横纵坐标是平均速度s和效果IR，图状表的感觉。
\section{Introduction}
% Image inpainting是计算机视觉中的一个重要任务，其旨在根据用户的需求为图像中的mask区域重绘合理的内容，同时保持背景部分与原图的一致性。随着text-to-image（t2i）模型的飞速发展，text-guided image inpainting任务越来越引起人们的重视，该任务区别于传统的inpainting任务的点在于，它主要基于用户输入的text prompt而不是图像背景上下文来对masj区域进行合理的内容填充。由于近期diffusion model在图像生成领域的突出表现，stable diffusion inpainting（SDI）及其衍生工作一直在该领域占据主导地位。
Image inpainting~\cite{li2022misf, deJorge_2024_ECCV, Liu_2022_CVPR, Dong_2022_CVPR} aims at filling masked regions and keeping harmonious with context. With the rapid development of text-to-image (T2I) generation~\cite{esser2024scaling, saharia2022photorealistic, sun2024uniavatar, liu2025moee}, text-guided image inpainting has gained significant attention. This task relies more on text prompts rather than solely on image context, with approaches like Stable Diffusion Inpainting (SDI)~\cite{rombach2022high,ho2020denoising, song2020score} leading the field.

\begin{figure}[t]
  \centering
   \includegraphics[width=1\linewidth]{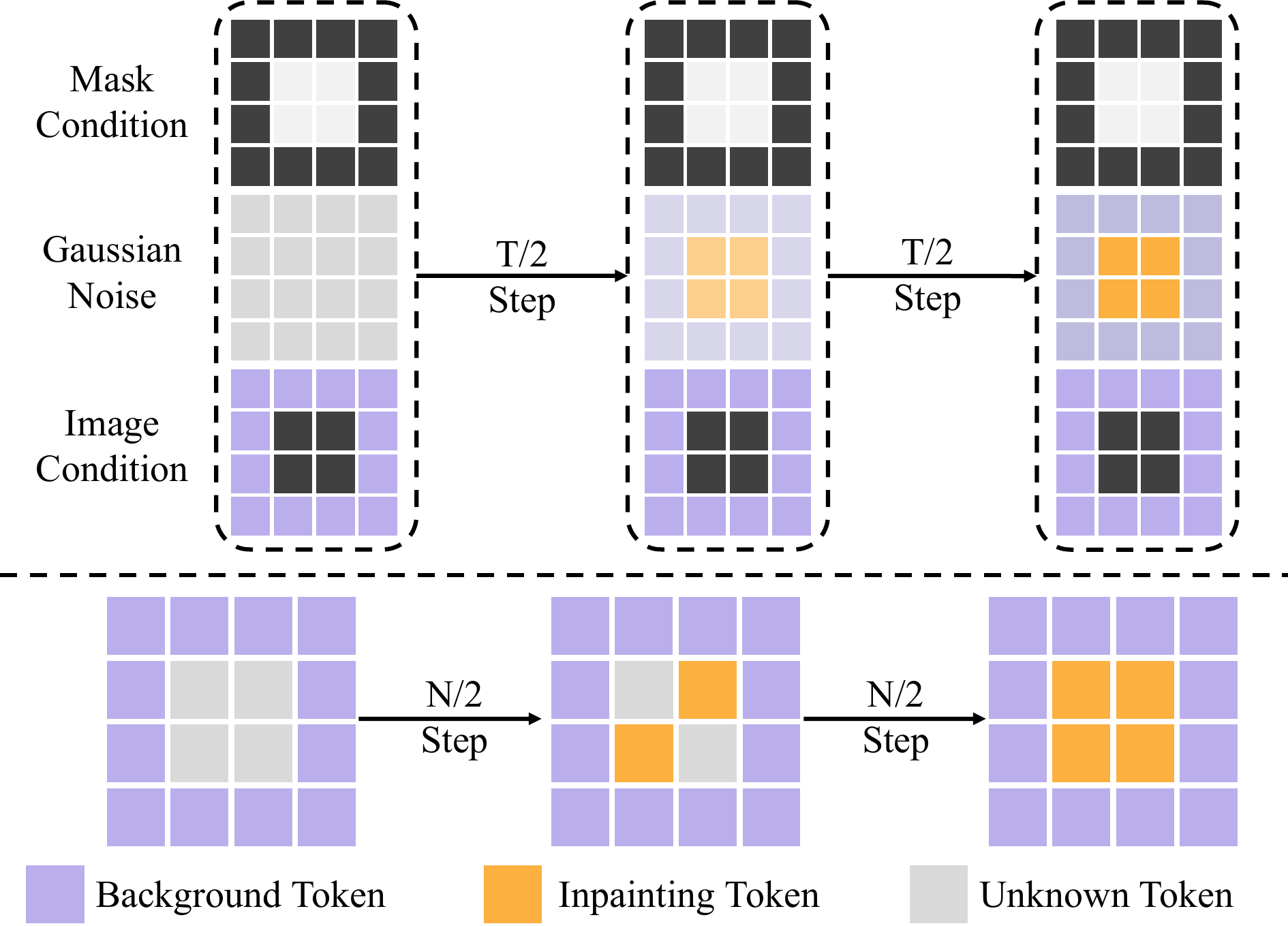}
   \vspace{-1.5em}
   \caption{Comparison of the inpainting process of diffusion models (above) and our MAR-based method (below).}
   \vspace{-2em}
   \label{fig:introduction}
\end{figure}

% SDI模型是使用stable diffusion文生图模型在包含text prompt的inpainting数据集上再次微调而来，因此继承了一定程度的生成质量性能。然而由于其微调的数据集采用random mask的策略，导致训练过程中需要根据text prompt条件填补的mask区域形状随机，这些mask往往不是一个实际物体的形状。同时由于mask的位置随机，可能导致重绘区域与背景物体重叠。这些训练数据上的问题限制了SDI模型的性能，带来了低质量及text prompt不符合的重绘内容产生的风险。一些Training-free的方法注意到了这些问题，并且基于SDI进行改进，例如HD-Painter通过重新加权注意力分数来改进采样策略，FreeCond则通过过滤低频分量来权衡背景信息和text信息的关注度，这些方法虽然在一定程度上有所改善，但难以真正的弥补数据问题带来的问题。还有一些方法尝试利用分割数据集对stable diffusion进行微调，然而分割数据集中mask标签的噪音及单调的文本描述也限制了训练后的模型的生成质量及指令跟随能力。更进一步的，diffusion model从纯噪声中对整幅图像进行去噪生成这一天然的生成特性破坏了背景部分的一致性。即使有些方法使用了blend手段对生成后的图片进行处理，也很难解决生成过程中背景部分的光照，色差等变化，导致拼接的边缘部分存在明显差异，并且这一操作本身也增添了繁琐的步骤和计算开销。
However, the diffusion-based text-guided inpainting faces potential challenges. The property of diffusion models that denoises the entire image in the latent space, causing the structure of the inpainting region to be guided more by context rather than text prompts~\cite{manukyan2023hd,hsiao2024freecond}. \textit{This leads to the inpainting content either being poorly aligned with the prompt or lacking harmony due to the conflict with context, resulting in low visual quality.} Some methods like HD-Painter~\cite{manukyan2023hd} and FreeCond~\cite{hsiao2024freecond} have noticed these issues and made training-free improvements based on SDI.
Other methods like BrushNet, PowerPaint~\cite{powerpaint,ju2024brushnet}, attempt to fine-tune stable diffusion (SD) while adding more powerful guidance. Although these methods show some improvement in text alignment, the quality of inpainting content is still unsatisfactory.
Furthermore, \textit{the generative property of denoising the entire image also inherently disrupts the consistency of the background}, as shown in Figure~\ref{fig:introduction}. 
And simple blending operations~\cite{ju2024brushnet} hardly resolve issues such as lighting inconsistencies or color mismatches, leading to noticeable seams.

% 近期，Autoregressive（AR）模型在生成领域受到了越来越多的关注。在生成过程中，传统AR模型首先在空间上将整个图像分成若干个token，然后按照扫描的顺序根据已知部分的token预测未知部分的token，这断绝了其在image inpainting任务中应用的可能性，其mask位置往往是随机的。然而最近AR模型出现了几种新的变体，在这其中，Mask Autoregressive（MAR）模型，由于其能够随机生成任意位置的图像token，使得其在image inpainting任务中的应用成为可能。更进一步的，MAR可以根据已知的图像区域的token来预测生成任意未知图像区域的token，因此其在生成过程中不会改变已知的背景图像token，保证了背景部分的一致性，因此相比于diffusion模型，MAR模型天然的更适合image inpainting任务。同时，与需要在特定inpainting数据集上微调的SDI模型不同，基于MAR的文生图模型天然的具有完成text-guide inpainting任务的潜力，这是因为其根据text prompt在mask区域中生成相关内容的这一意图与其在训练过程中根据text prompt生成图像的这一目标具有相当大的重叠，使得其遵循prompt生成的inpainting内容的生成质量不会被限制，具有极大的潜力。因此，我们的工作就是挖掘基于MAR的文生图模型在text-guided image inpainting任务上的潜力。
Recently, Autoregressive (AR) models~\cite{vaswani2017attention, radford2018improving,sun2024autoregressive, team2024chameleon} have garnered increasing attention in T2I generation. In AR models, each token in latent space corresponds to a part of the image spatially, and unknown tokens are directly predicted based on known tokens at each step during the generation process. Compared to the joint denoising of all tokens in diffusion, the property of AR increases the structural flexibility of sub-regions, leading to better controllability of local content generation. But traditional AR models predict tokens through a raster-order, limiting their applicability in image inpainting, where the mask positions are typically random. Recently, variants of AR models~\cite{tian2024visual, pang2025randar, yu2025frequency, li2024autoregressive} have emerged, among which the mask autoregressive (MAR)~\cite{li2024autoregressive} stands out. Unlike traditional AR models, MAR generates image tokens at arbitrary locations, making it suitable for image inpainting. In addition, MAR inherits the property of AR that generates tokens at each step, not only making it have strong local controllability for inpainting tokens, but also keeping known background tokens unchanged during generation. As shown in Figure~\ref{fig:introduction}, MAR is naturally suitable for inpainting. 
Therefore, our work focuses on applying recent T2I MAR model NOVA~\cite{deng2024autoregressive} with an encoder-decoder architecture to text-guided image inpainting.
% have the inherent potential to perform text-guided inpainting, due to the goal of generating inpainting content based on the text prompt overlaps significantly with its training objective of generating images based on text description. This alignment ensures that the quality of the inpainting content is not limited, offering substantial potential for high-quality inpainting generation. Therefore, our work focuses on exploring the potential of MAR models in the text-guided image inpainting task.

% 我们从在最近使用encoder-decoder结构的T2I MAR模型上一个简单的做法开始，首先我们将用于inpainting内容的text作为prompt输入，同时我们将潜在空间中对应于mask区域的图像token去除，然而输出的结果无视了text prompt，仅参考了图像上下文。之后，我们再次将text作为prompt输入，与之前不同的是，这次MAR将无视其余背景token条件，结果虽然遵循了text prompt，但是生成内容却与背景极其不和谐。为了进一步确定两次重绘失败的原因，我们分别对这两次实验decoder中的text-image交叉注意力图以及inpainting区域的自注意力图进行了可视化，结果表明在第一个分支中，两种注意力图的关注区域分散到了图像的背景部分，而第二个分支中，两种注意力图的关注区域都集中在了inpainting区域内。因此我们认为第一种做法中text token的信息在encoder交互过程中被背景token的信息所淹没覆盖。而在第二种做法中因为缺少背景token在encoder中的交互，所以text token完全保持了原始的text信息，但因此生成的token却背景上下文极度不和谐。
Our study starts by setting both text prompt and background tokens (T\&B) as the input to generate the masked region. However, the inpainting content ignores the text prompt and relies solely on the context, as shown in Figure~\ref{fig:two_branch}(a). Then, we only use the text prompt as input and disregard all background tokens (T-only). Although the inpainting content follows the prompt this time, it is highly disharmonious with the surrounding context. To further investigate the reasons for these failures, we visualize two types of self-attention maps in the decoder stage for both approaches, as shown in Figure~\ref{fig:two_branch}(b). The results reveal that in the T\&B case, the attention scores of both types are dispersed to background regions, while in the T-only case, the attention scores are concentrated within the inpainting region. Therefore, we infer that in the T\&B approach, the semantic information of text tokens is overwhelmed by the context information of background tokens during encoder interaction. In the T-only approach, due to the lack of background tokens, the text tokens fully retain their semantic information, but the generated tokens are highly disharmonious with image context.

% 为了解决上述问题，我们提出了Token Painter，一种基于MAR的training-free text-guided image inpainting方法，其分别在encoder和decoder两个阶段对T2I MAR模型进行了针对text inpainting任务的改进。具体而言，Token Painter在encoder阶段提出了Dual-Stream Encoder Information Fusion (DEIF)模块，该模块旨在获得一个优异的引导特征，也就是经过encoder交互后的text token特征。使之在其引导下，模型在decoder过程中能够生成遵循text prompt同时与图像上下文和谐的inpainting内容。该模块利用首先将上述两种简单方法分支中两种粗糙的引导特征在统计量上进行对齐，然后在频域上进行了语义和背景信息的分离及再融合，使之既包含决定语义内容的低频分量同时也包含决定背景内容的高频分量。而在decoder阶段，为了更好的使inpainting区域更好关注引导特征及区域内部token的信息，我们精心设计了Adaptive Decoder Attention Score Enhancing (ADAE)模块。它不仅提高inpainting区域对引导特征的关注度，还根据MAR模型的生成过程，对inpainting区域内的自注意力分数进行动态调整。通过结合DEIF和ADAE，我们的Training-free方法在更少的参数量级上甚至超过了基于微调模型SDI进行改进的现有的SOTA方法。
To address the above issues, we propose Token Painter, a training-free text-guided image inpainting method based on MAR. We design it at both encoder and decoder stages. At the encoder stage, we present the Dual-Stream Encoder Information Fusion (DEIF) module, which aims to produce novel guidance tokens, \textit{i.e.}, the updated text tokens after encoder interaction. With those guidance tokens, the model can generate inpainting content that follows the text prompt and keeps in harmony with the image context. The module first aligns the two rough guidance tokens from T\&B and T-only statistically, and then fuses them in the frequency domain via a modified Gaussian function. At the decoder stage, we design the Adaptive Decoder Attention Score Enhancing (ADAE) module to further improve the prompt detail alignment and the content visual quality. This module enhances the attention of inpainting tokens to guidance tokens adaptively, and strengthens the attention interaction within the inpainting region based on the property of MAR generation. Token Painter outperforms existing state-of-the-art (SOTA) methods, including those models fine-tuned on inpainting datasets. Contributions are summarized as follows:

% 据我们所知，我们是第一个基于MAR进行text-guided image inpaiting任务的工作。同时我们深入研究分析了MAR在text inpainting任务中的机制，揭示了text token及图像上下文token在生成过程中的交互作用及影响。
% 我们引入了Dual-Stream Encoder Information Fusion (DEIF) module来得到能够同时生成遵循text prompt且与上下文和谐的inpainting内容的引导特征。
% 为了进一步增强对text对齐能力和inpainting能让的生成质量，我们引入了Adaptive Decoder Attention Score Enhancing (ADAE)模块，对交叉注意力和自注意力分数进行动态调整。
% 我们的方法Token Painter，虽然是Training-free且拥有更少的参数，却在几乎所有指标上中超越了过往的方法包括SOTA，即使他们全部都基于在inpaiting数据集上微调过的模型。
\begin{itemize}
    \item We improve the T2I MAR model specifically for the text-guided image inpainting task, and conduct a detailed analysis of MAR text-guided inpainting process, revealing the interactions between text tokens and background tokens, as well as their impact on the inpainting content.
    \item We introduce the Dual-Stream Encoder Information Fusion (DEIF) module to obtain novel guidance tokens that guide MAR to generate inpainting content that follows the text prompt while keeping harmonious with context.
    \item To further enhance the alignment of prompt details and the visual quality of inpainting, we introduce the Adaptive Decoder Attention Score Enhancing (ADAE) module, which adaptively enhances the attention scores.
    \item Token Painter is a training-free method based on the T2I MAR model, yet it outperforms previous methods across nearly all metrics, including SOTA, even though they are based on models fine-tuned on inpainting datasets.
\end{itemize}

\section{Related Work}
% Traditional image inpainting.
% 随着深度学习技术的快速发展，image inpainting这一任务在变分自编码器和生成对抗网络架构的基础上涌现出了一些经典的方法。这些方法在训练过程中利用随机生成的mask对来源于现实场景的数据集图片进行遮掩作为输入。这些传统的模型虽然能够生成与图像上下文和谐的inpainting内容，但却难以针对特定主题产生全新的内容。
% Text-guided image inpainting.
% 近些年来，各种多模态任务的出现使得text-guided image inpainting任务越来越受到人们的关注。随着diffusion模型在视觉生成中大放异彩，许多基于diffusion的工作也随之出现。一些training-free的方法在stable diffusion inpainting的基础上进行改进，而其他一些方法则利用图像分割数据集对stable diffusion模型进行微调。然而低质量的mask标签及text描述限制了这些工作的生成质量和指令遵循能力，同时diffusion的生成特点也损害了背景一致性。
% add marmeissonic
\subsection{Text-guided Image Inpainting}
% Though those models are able to generate inpainting content that is harmonious with the image context, they still struggle to produce novel content.

% \noindent\textbf{Text-guided Image Inpainting.} In recent years, the emergence of T2I generation models has led to increased attention on text-guided image inpainting. With the success of diffusion models in visual generation, many diffusion-based works
% \noindent\textbf{Text-guided Image Inpainting.} 
% In recent years, the development of T2I generation models has led to increased attention on text-guided image inpainting. Many diffusion-based text-guided image inpainting works~\cite{nichol2021glide, bld, avrahami2022blended, wang2023imagen} have emerged. Some training-free~\cite{manukyan2023hd,hsiao2024freecond} methods improve upon SDI, while others~\cite{powerpaint,ju2024brushnet} fine-tune the SD using segmentation datasets. However, the property of diffusion limits the visual quality and prompt alignment of these methods, and also disrupts background consistency.
In recent years, many diffusion-based text-guided image inpainting works~\cite{nichol2021glide, bld, avrahami2022blended, wang2023imagen} have emerged. However, the property of diffusion limits visual quality and prompt alignment of these methods, and also disrupts background consistency. There are some T2I MAR models~\cite{chang2023muse, bai2024meissonic} that involve this task, but their naive approach and model architecture including feature compression layers lead to low-quality inpainting and inconsistent background.

%基于diffusion和AR的文生图模型
%文生图任务在过去几年一直被基于diffusion的方法所主导。得益于连续的空间采样建模，其生成的视觉内容往往更具多样性，纹理更加细腻。然而，随着AR模型逐渐进入视觉生成领域，人们发现基于AR的生成模型往往更能更好的遵循text指令生成对应的空间结构，这是由于潜在空间中的图像token准确的在空间结构中对应着生成图像。
%基于MAR的文生图模型
% 经典的AR模型是根据扫描的顺序进行图像的生成，然而这种方式与直觉相冲突。因此AR模型在视觉领域产生了多种变体，其中MAR通过生成任意位置复数token的方式，在视觉结构合理性和图像质量上尤为突出。基于MAR的T2I模型也由于其优异的生成质量和text细节对齐能力为人们所关注。该模型采用了encoder-decoder结构，通过将含有语义信息的固定长度text token与初始化的image token拼接作为输入。在encoder阶段将text token和已生成的图像token进行交互，而在decoder对未知的图像token进行预测。由于其图像token生成的灵活性和良好的文生图能力，我们在其基础上进行text-guided image inpainting任务的研究。
\subsection{Text-to-Image Generation}
\textbf{Diffusion and Autoregressive Models.} Text-to-image generation, has been dominated by diffusion-based methods~\cite{betker2023improving, chen2023pixart, esser2024scaling} in recent years.
% Thanks to continuous spatial sampling strategy, the generated visual content is often more diverse and has finer textures. 
However, as AR models~\cite{fan2410fluid, sun2024autoregressive, yu2022scaling} gradually enter the field of visual generation, it has been found that AR-based T2I models tend to follow text prompts better to generate the corresponding spatial structure. This is due to the higher independence between different image tokens in the AR models. 

\noindent\textbf{Mask Autoregressive Model.} The classic AR model generates image tokens according to the raster-order, which conflicts with intuition. As a result, variants of AR models have emerged, among which MAR models~\cite{li2024autoregressive,chang2022maskgit} stands out for its ability to generate tokens at arbitrary positions.
% excelling in both visual structure and image quality. 
% The MAR-based T2I model~\cite{deng2024autoregressive} has gained attention due to its superior generation quality and alignment with text details. This model uses an encoder-decoder structure, where fixed-length text tokens containing semantic information are concatenated with initialized image tokens as input. During the encoder stage, the text tokens interact with the generated image tokens, and in the decoder, unknown image tokens are predicted. Due to its flexibility in generating local image tokens and strong text-to-image ability, we conduct research on the text-guided image inpainting task based on this model.
% The T2I MAR models~\cite{deng2024autoregressive, chang2023muse, bai2024meissonic} has also gained attention due to its superior generation quality and alignment with text details. We conduct research on this model to achieve text-guided image inpainting task.
The T2I MAR models~\cite{deng2024autoregressive, chang2023muse, bai2024meissonic} has also gained attention due to their superior text alignment. Considering visual generation quality and background region consistency, we choose the T2I MAR model NOVA\cite{deng2024autoregressive} as base model to achieve text-guided image inpainting task.

\section{Analysis of MAR Model for Inpainting}
% 定义两种极端，分别介绍
% 首先，我们来回顾一下具有encoder-decoder结构的T2I MAR模型的主要生成过程。给定一个text prompt，模型中的text encoder部分将其转换为一个固定长度的text token特征F。然后模型将初始化一组完全未知图像token并将其分为若干组，每组都包含若干个图像token。接着模型通过text token特征和已生成的图像token以组为单位预测未知的图像token，如下所示：
%公式
%其中x是图像。在这个过程中，text token特征和已生成的图像token首先在encoder中进行信息交互传递，得到交互后的text token特征，也即guidance tokens和交互后的图像特征。decoder则根据这些交互后的特征来预测下一组图像token。
%为了将T2I MAR模型应用到text-guided image inpainting任务上，我们简单的将生成范式进行了更改，将图像token中的inpainting部分变为未知token，保留背景token作为已知token，调整过的简单范式如下所示：
%公式
%其中x是图像。然而我们发现这种简单的范式生成的结果却完全不遵循给定的text prompt，而仅参考图像上下文信息进行inpainting。为了使得模型生成的inpainting内容能够与text进行对齐，我们将已知的图像token信息完全屏蔽，范式更改如下：
%公式
%其中x是图像。这次模型虽然能够生成对应text prompt的内容，但是inpainting部分却与图像上下文完全不和谐，使得视觉效果呈现出很强的割裂感。

\subsection{Vanilla Solutions for MAR-based Inpainting}
We firstly revisit the main generation process of the T2I MAR model NOVA~\cite{deng2024autoregressive} with an encoder-decoder architecture. Given a text prompt, the text encoder converts it into the fixed-length text tokens $T \in \mathbb{R}^{L\times D}$. The model then initializes a group of unknown image tokens $I \in \mathbb{R}^{H\times W\times D}$, and divides them into $V$ sets $\{S^{1}, S^{2}, ..., S^{V}\}$. Those sets are predicted in order based on text tokens and known image tokens. This paradigm is written as:
\begin{equation}
p(S^1,...,S^V)=\prod_{v}^{V}p(S^v\mid T, S^1,...,S^{v-1}),
\end{equation}
where $S^v$ is a set to be predicted at $v$-th step, with $\cup _vS^v=I$. During this process, the text tokens $T$ and the known image tokens interact within the encoder, resulting in the updated text tokens, i.e., guidance tokens $T_g$. The decoder then predicts the next set of image tokens mainly based on them.

\begin{figure}[t]
  \centering
   \includegraphics[width=1\linewidth]{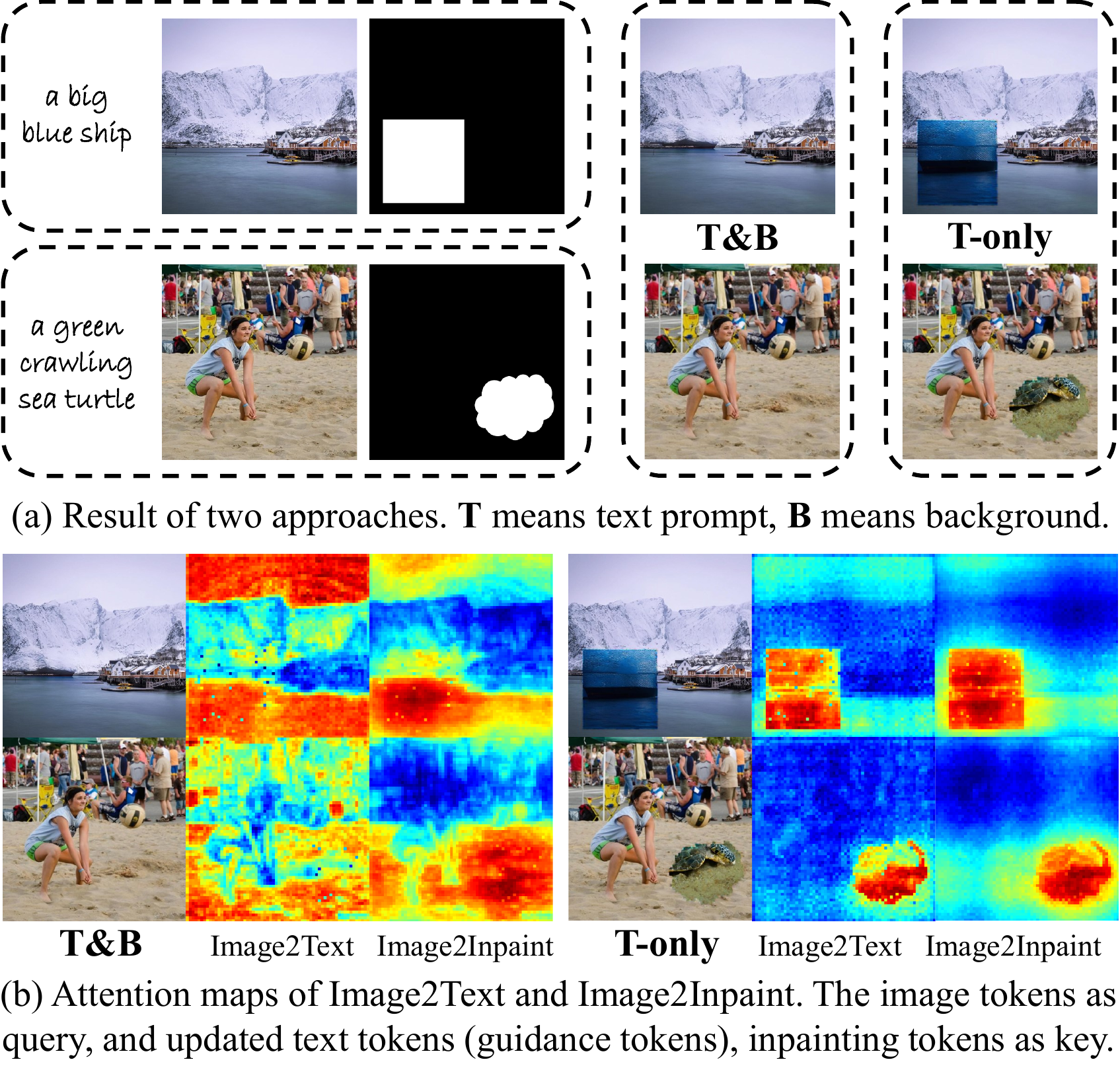}
   \vspace{-1.5em}
   \caption{Comparison of vallia T\&B and T-only approaches.}
   \vspace{-1.5em}
   \label{fig:two_branch}
\end{figure}

To apply this paradigm to the text-guided image inpainting task, we make a simple modification. We label the set of inpainting tokens $I_{p}\in \mathbb{R}^{N\times D}$ as unknown, while keeping the background tokens $I_{b}\in \mathbb{R}^{M\times D}$ as predicted tokens, where $M+N=HW$. The modified paradigm is as follows:
\begin{equation}
p(S^1,...,S^k)=\prod_{k}^{K}p(S^k\mid T, I_{b}, S^1,...,S^{k-1}),
\end{equation}
where $S^k$ is a set to be predicted at $k$-th step, with $\cup _kS^k=I_p$. However, we find that the inpainting region generated by this approach (T\&B) does not follow the prompt at all, as shown in Figure~\ref{fig:two_branch}(a). The content seems to only rely on the image context. Then we completely mask the background tokens (T-only). The modified paradigm is as follows:
\begin{equation}
p(S^1,...,S^k)=\prod_{k}^{K}p(S^k\mid T, S^1,...,S^{k-1}).
\end{equation}
As shown in Figure~\ref{fig:two_branch}(a), though the inpainting region follows the prompt now, it is disharmonious with context.

\begin{figure*}[t]
    \centerline{\includegraphics[width=1.0\linewidth]{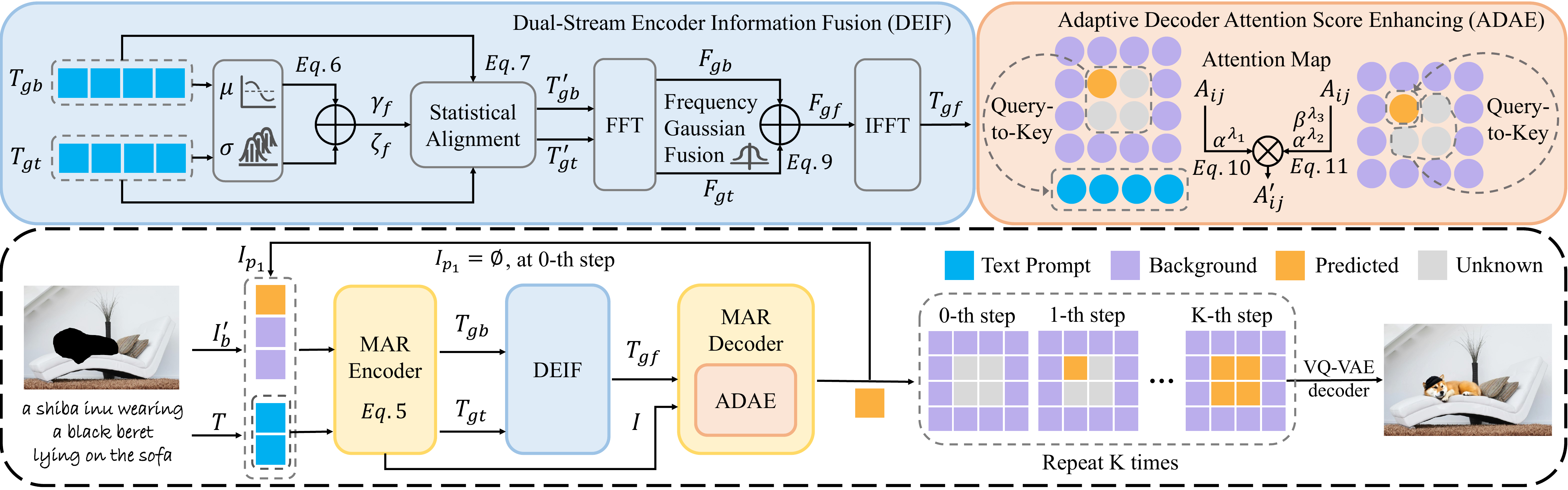}}
    \vspace{-0.5em}
    \caption{Overview of Token Painter, which includes the DEIF at encoder stage and the ADAE at decoder stage. DEIF produces novel guidance tokens $T_{gf}$ that contain both text and context information through information fusion in frequency domain. ADAE enhances two parts of attention map $A$ to further improve prompt detail alignment and content visual quality.} 
    \label{fig:mainmethod}
    \vspace{-1.5em}
\end{figure*}

% 两种注意力图，统计分量
% 为了更清晰的了解上述两个分支中guidance tokens对inpainting内容造成的影响，我们分别对上述两种简单范式中的两类自注意力图进行可视化。第一类为guidance tokens作为keys值对image tokens作为queries的注意力图，第二类为inpainting tokens作为keys对image tokens作为queries值的注意力图。我们分别将这两类注意力图沿着key值维度取均值并重新resize为HxW形状，如图所示。可以看到，第一种简单范式下，guidance tokens的注意力值分散到了全图中，特别是图像中的背景部分。同时inpainting token不仅与自身密切相关，对周围的图像上下文也有较高的相似度。第二种范式下，两种token的注意力值都严格的限制在了inpainting区域。因此我们可以提出假设，在第一种情况下，当背景token全部可见与text token一同进入encoder进行交互时，产生的guidance tokens中的text语义信息被背景信息所淹没，因此在其引导下生成的inpainting区域自然的偏向于图像上下文内容。而当背景token被完全屏蔽时，由于缺乏上下文信息进行交互，guidance中仅有text prompt的语义信息而无背景信息，这一特点则导致了生成内容了与上下文极度不和谐。
%为了验证这一猜想
\subsection{The Mechanism behind MAR Inpainting}
To better understand the impact of the guidance tokens in T\&B and T-only approaches, we visualize two types of attention maps for both, as shown in Figure~\ref{fig:two_branch}(b).
% The first map shows the attention between image tokens $I$ (queries) and guidance tokens (keys) $T_g$, while the second map shows the attention between image tokens $I$ (queries) and inpainting tokens $I_p$ (keys). We average these attention maps along the key dimension and resize them to the shape of $H\times W$. 
In the T\&B case, the attention scores of the guidance tokens are distributed across the entire image, especially in the background. Meanwhile, the inpainting tokens show high similarity with both themselves and the surrounding image context. In the T-only case, the attention scores for both token types are strictly limited to the inpainting region.
Based on these observations, we propose that in the T\&B case, where background tokens are fully visible and interact with text tokens during encoder stage, the semantic information in the text tokens is overwhelmed by the context information. As a result, the inpainting region generated under these guidance tokens tends to align with image context. Conversely, when background tokens are masked, the lack of context interaction means the guidance tokens contain only semantic information, leading to the disharmonious inpainting content.

\section{Token Painter}
% 我们方法的概述在图中呈现。Token Painter主要在两个阶段对T2I MAR模型NOVA进行调整以适应text-guided image inpainting模型，即encoder阶段和decoder阶段，分别对应于Dual-Stream Encoder Information Fusion (DEIF) module 和 Adaptive Decoder Attention Score Enhancing (ADAE) module。
% Text-guided Image inpainting任务主要有三部分输入：text prompt，image和mask。text prompt经由text encoder编码为固定长度的text tokens。对于image和mask，我们首先使用VQ-VAE将image编码进latent space，然后将mask使用最近邻下采样算法resize至相同尺寸，并将其与latent image相乘得到去除inpainting部分的latent image。接着我们分别将text token加上已预测的图像token（包含部分环绕inpainting区域的背景token）和单纯的text token送入encoder，得到交互后的双支路guidance tokens。将这两种分别带有背景信息和语义信息的guidance tokens送入DEIF模块。该模块首先在统计量上将二者进行对齐，然后使用FFT将其转换到频域。在频域空间使用调整后的高斯函数插值组合二者的低频语义结构信息和高频风格信息，再经过IFFT变换得到新的guidance tokens。
% 接着guidance tokens将被送入decoder中指导生成inpianting区域的内容。在此过程中，ADAE模块根据总计生成的token数量对自注意力图的两个特定部分进行自适应增强。第一是以guidance tokens作为keys且以inpainting token作为queries。第二是以已预测的inpainting token作为keys且以未知的inpainting token作为queries的部分。其中第二部分还根据MAR的特性实行了动态增强系数。以上过程重复K次，直至生成所有inpainting token后，与背景token进行拼接。然后使用VQ-VAE解码得到最终的inpainting 图像。

\subsection{Overview of Our Method}
Overview of Token Painter is presented in the Figure~\ref{fig:mainmethod}. The text prompt is converted into fixed-length text tokens $T$. Then a VQ-VAE encode the image to $I$. The mask is downsampled to $M$ and multiplied with $I$ to produce $I_{b}$ and unknown $I_{p}$. Next, we input those tokens to the MAR encoder in the way of T\&B and T- only to get $T_{gb}$ and $T_{gt}$, and DEIF module fuse them to get the novel guidance tokens $T_{gf}$. It is then fed into the decoder to guide the generation of the inpainting region, with ADAE module adaptively enhancing two parts of the attention map $A$. After repeating $K$ times, we decode latent tokens by VQ-VAE to obtain final image.

\tabcolsep=0.17cm
\begin{table*}[h]
\small
    \centering
    \begin{tabular}{p{0.1\linewidth}ccccccccccc}%
    \toprule
    Dataset & Methods & IR$_{\times10}$~$\uparrow$ & PS$_{\times10^{2}}$~$\uparrow$ & HPS$_{\times10^{2}}$~$\uparrow$ & AS~$\uparrow$ & PSNR~$\uparrow$ & LPIPS$_{\times10^{3}}$~$\downarrow$ & SSIM$_{\times10}$~$\uparrow$ & CLIP-S~$\uparrow$ \\
    \midrule
    \multirow{7}{*}{EditBench} & SDI & -8.72 & 41.90 & 21.39 & 3.79 & \underline{24.47} & 28.70 & 8.50 & 24.39 \\
    & HD-Painter & -6.16 & 48.64 & 22.73 & 3.79 & 24.08 & \underline{25.64} & 8.46 & 25.62 \\
    & PowerPaint & \underline{-5.11} & \underline{50.13} & \underline{22.94} & 3.86 & 24.45 & 25.69 & \underline{8.69} & 26.05 \\
    & BrushNet & -7.50 & 46.37 & 22.45 & 3.80 & 22.71 & 30.73 & 8.31 & \textbf{26.14} \\
    & FreeCond & -7.13 & 47.16 & 22.90 & 3.88 & 22.38 & 36.88 & 8.13 & 25.15 \\
    & Meissonic & -6.66 & 47.02 & 22.70 & \underline{3.90} & 22.24 & 35.66 & 8.31 & 24.31 

    \\\cmidrule{2-10}
    & Token Painter (Ours) & \textbf{-2.49} & \textbf{55.37} & \textbf{23.00} & \textbf{3.92} & \textbf{28.03} & \textbf{24.92} & \textbf{9.41} & \underline{26.06} \\

    \midrule
    \multirow{7}{*}{BrushBench} & SDI & 11.89 & 41.30 & 27.20 & 4.20 & 22.82 & \underline{43.54} & 7.68 & 14.44\\
    & HD-Painter & 10.81 & 37.25 & 26.96 & 4.21 & 20.98 & 49.34 & 7.61 & 14.37 \\
    & PowerPaint & 11.96 & 42.46 & \underline{27.65} & 4.13 & \underline{23.43} & 51.46 & \underline{7.96} & \underline{14.45} \\
    & BrushNet & 12.42 & 41.15 & 27.51 & \textbf{4.25} & 21.84 & 47.87 & 7.59 & 14.44\\
    & FreeCond & 11.97 & 40.44 & 27.61 & 4.21 & 21.49 & 50.03 & 7.43 & 14.40 \\
    & Meissonic & \underline{12.51} & \underline{44.34} & 27.56 & 4.21 & 22.35 & 64.81 & 7.72 & 14.44 

    \\\cmidrule{2-10}
    & Token Painter (Ours)  & \textbf{13.01} & \textbf{47.90} & \textbf{28.36} & \underline{4.22} & \textbf{26.39} & \textbf{42.27} & \textbf{8.78} & \textbf{14.46} \\
    \bottomrule
    \end{tabular}
    \vspace{-0.5em}
    \caption{Quantitative results on the EditBench and BrushBench datasets. The \textbf{best results} and the \underline{second best results} are marked in bold and underline, respectively. Results are all re-evaluated based on the released code, models and setting. }
    \vspace{-2em}
    \label{tab:quantitative_comparison}
\end{table*}

\subsection{Dual-Stream Encoder Information Fusion}
% 为了得到同时包含语义信息和背景信息的guidance tokens，我们需要对第三部分提到的两种简单方法下的guidance tokens进行融合。这里我们分别将两种text token在encoder中交互得到的guidance tokens（带有背景token和不带有背景token）定义为Tgb和Tgt。在实际应用过程中，我们发现Tgt仅需与inpainting区域周围的部分背景token交互，即可很好的包含上下文信息。因此，这里我们使用膨胀后的mask减去原始mask后，其中对应的surrounding背景token，选定过程如下所示：
%公式
%其中x是图像。接着，我们分别将上述surrounding背景token加上text token以及单独text token二者分别送入MAR的encoder中进行交互，然后得到两种简单的guidance tokens，即Tgb和Tgt。过程如下所示：
%公式
%其中x是图像。
\textbf{Dual-Stream Guidance Tokens.}
To obtain a guidance tokens that includes both semantic and context information, we need to fuse the two rough guidance tokens, $T_{gb} \in \mathbb{R}^{L\times D}$ and $T_{gt} \in \mathbb{R}^{L\times D}$, from the T\&B and T-only approaches mentioned in Section~3. In practical application, we find that $T_{gb}$ only needs to interact with the background tokens surrounding the inpainting region, and it effectively captures contextual information. Therefore, we use the mask $M\in \mathbb{R}^{H\times W\times 1}$ and its dilated version $M_d$ to choose the background tokens for interaction. The process is as follows:
\begin{equation}
I'_{b}=(I\odot(M_{d}-M))[:p\cdot N, :],
% \in \mathbb{R}^{pN\times D},
\end{equation}
where $p$ is a proportionality coefficient and $N$ is the number of inpainting tokens. Next, we feed the text tokens with the known image tokens, as well as the text tokens alone, into the MAR encoder for interaction. The process is as follows:
\begin{equation}
\begin{aligned}
T_{gb}&=ME(Concat(T,I'_{b},I_{p_1}))[:L, :] \in \mathbb{R}^{L\times D}, \\
T_{gt}&=ME(T)[:,:] \in \mathbb{R}^{L\times D},
\end{aligned}
\end{equation}
where $Concat(\cdot)$ means the operation of concatenation, and $ME(\cdot)$ represents the interaction of encoder in MAR.

%为了更好的在频域中将二者进行融合，我们首先需要将其在统计量上进行对齐，也就是对双方进行标准化后重新将二者shift至某一共同分布。由于在对齐时，我们需要同时兼顾每个实例的语义信息和背景信息的统计特性，因此我们shift的均值方差为在每次inference中进行自适应变化，由下式得到：
%公式
%其中x是图像。在得到指定的shift的均值方差后，我们将两种guidance tokens进行对齐，如下式所示：
%公式
%其中x是图像。
\noindent\textbf{Adaptive Statistical Alignment.}
To better fuse $T_{gb}$ and $T_{gt}$ in the frequency domain, we first need to align them statistically in the space domain, which involves normalizing both $T_{gb}$ and $T_{gt}$, then shifting them to a common distribution. Since the alignment process must account for the statistics of both semantic and context information of each instance, the mean $\gamma_{f}$ and variance $\zeta_{f}$ for the shift are adaptively obtained in each instance, as described by the following equation:
\begin{equation}
\begin{aligned}
\gamma_{f}&=a\cdot\mu(T_{gb})+(1-a)\cdot\mu(T_{gt}), \\
\zeta_{f}&=a\cdot\sigma(T_{gb})+(1-a)\cdot\sigma(T_{gt}),
\end{aligned}
\end{equation}
where $\mu(\cdot)$ represents the mean of tokens along $L$ dimension, and $\sigma(\cdot)$ represents the variance. $a$ is a proportional coefficient. Then we align $T_{gb}$ and $T_{gt}$ as follows:
\begin{equation}
\begin{aligned}
T'_{gb}&=\gamma_f\cdot\left(\frac{T_{gb}-\mu(T_{gb})}{\sigma(T_{gb})}\right)+\zeta_f, \\
T'_{gt}&=\gamma_f\cdot\left(\frac{T_{gt}-\mu(T_{gt})}{\sigma(T_{gt})}\right)+\zeta_f.
\end{aligned}
\end{equation}

%在得到两种对齐过的guidance之后，我们需要将Tgb的背景风格信息与Tgt的语义结构信息进行融合得到新的引导特征。我们首先将二者通过FFT转换到频域并将其零频率分量移到频谱中心，也就是L/2处，得到Fgb和Fgt。接着我们使用一个调整后的高斯函数对二者平滑的进行插值组合得到Fgf，使得高频部分主要包含Tgb的背景风格信息，而低频部分主要包含Tgt的语义结构信息。然后对Fgf进行IFFT即可得到最终的指导特征来引导后续inpainting内容生成。整个过程如下所示：
%公式
%其中x是图像。
\noindent\textbf{Frequency Information Fusion.}
Inspired by~\cite{kwon2024aesfa, gao2024frequency}, we need to fuse the high-frequency context style information from $T'_{gb}$ with the low-frequency semantic structure information from $T'_{gt}$. We first transform them into the frequency domain using fast fourier transform (FFT) and shift their zero-frequency components to the center of the frequency spectrum, i.e., at $L/2$, to obtain  $F_{gb}\in\mathbb{R}^{L\times D}$ and $F_{gt}\in\mathbb{R}^{L\times D}$. Next, we use a modified Gaussian function to fuse the two frequency spectra, yielding $F_{gf}$. 
% This enables $F_{gf}$ to have the high-frequency context style information from $T_{gb}$ and low-frequency semantic structure information from $T_{gt}$. 
The entire process is illustrated below:
\begin{equation}
MG(l)=exp\left(-(\frac{|l-L/2|}{\varphi})^{\tau}\right),
\end{equation}
\begin{equation}
F_{gf}(l)=(1-MG(l))\cdot F_{gb}+MG(l)\cdot F_{gt},
\end{equation}
where $MG(l)$ is the value of the modified Gaussian function at position $l\in \{0,1,..., L-1\}$. $\varphi$ and $\tau$ are coefficients used to control the shape. Afterward, we shift the $F_{gf}$ to the original frequency spectrum, and apply the IFFT to it to obtain the novel guidance tokens $T_{gf}\in\mathbb{R}^{L\times D}$.

\begin{figure*}[t]
  \centering
   \includegraphics[width=1\linewidth]{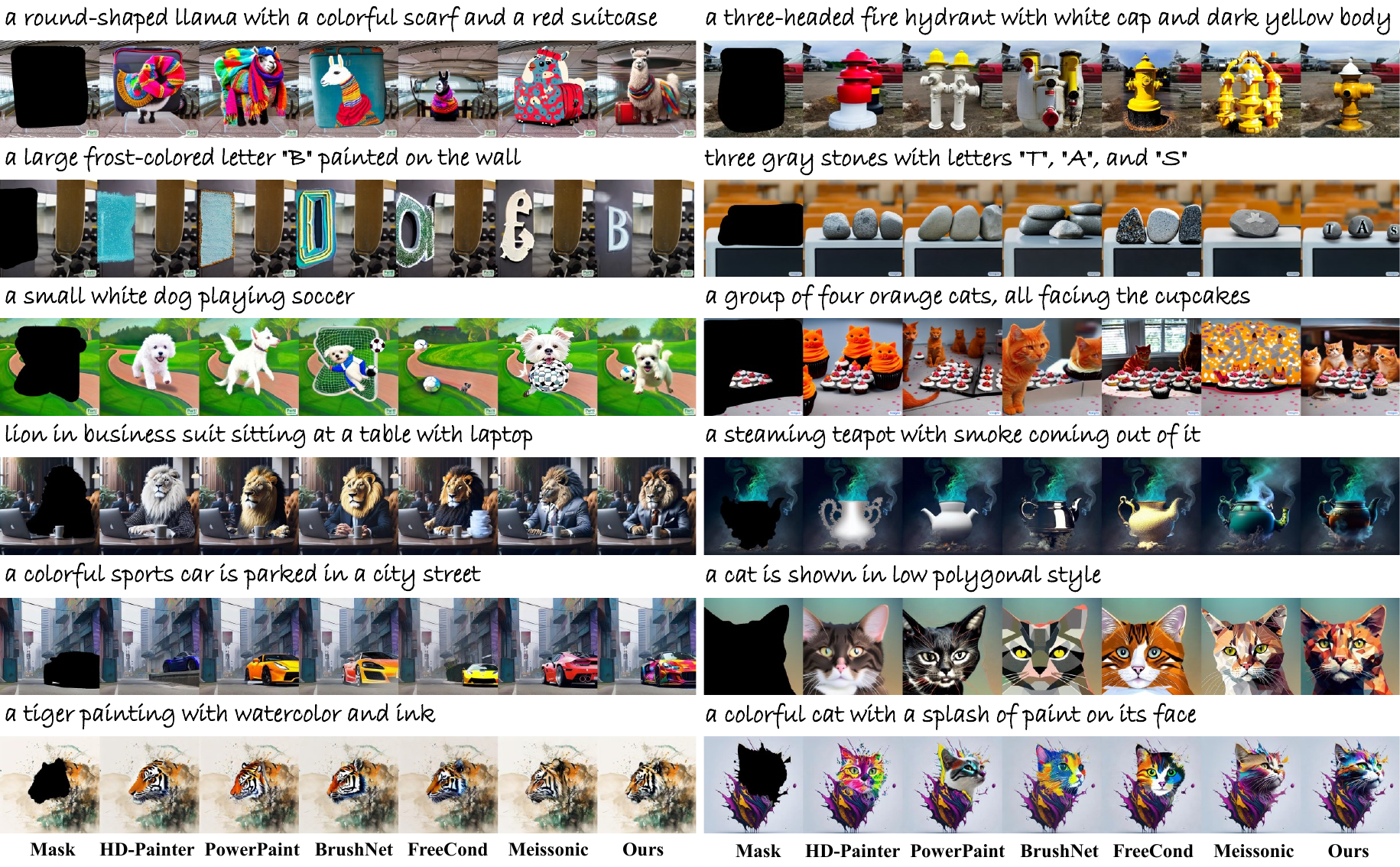}
   \vspace{-2em}
   \caption{Qualitative results of our Token Painter with previous text-guided inpainting methods. The first three rows of samples are from EditBench with loose masks, and the last three rows of samples are from BrushBench with tight masks.}
   \vspace{-1.5em}
   \label{fig:qualitative_comparison}
\end{figure*}

\subsection{Adaptive Decoder Attention Score Enhancing}
%将上述得到的新指导分量送入MAR的decoder，模型会在其指导下生成inpainting内容。为了进一步提高prompt细节忠诚度和视觉质量，我们对自注意力图中的两个部分进行了自适应动态增强。
%在进行增强前，我们需要确定对注意力分数的增强系数。由于不同实例下的inpainting区域大小不同，因此增强系数需要对应于inpainting token的数量。受到attention entropy的启发，我们认为该系数主要取决于training阶段与inference阶段生成的token数量的差异。因此我们将增强系数设定为随着inpainting token数量减小而增大的log函数，如下式所示，
%公式
%其中x是图像。
\textbf{Adaptive Enhancement Coefficient.}
To further enhance the prompt detail alignment and content visual quality, we apply adaptive enhancement to two parts of the attention map.
% Before enhancing, we need to determine the enhancement coefficient for the attention scores. 
Since the size of the inpainting region varies across instances, the enhancement coefficient must correspond to the number of inpainting tokens. Inspired by~\cite{jin2023training}, we propose that this coefficient primarily depends on the difference in the number of tokens generated during the training and inference stages. 
Therefore, we set this coefficient to follow a logarithmic function that increases as the number of inpainting tokens decreases, \textit{i.e.}, $\alpha=\mathrm{log}_NHW$.

% 在MAR的decoder中，guidance tokens和所有image tokens拼接在一起作为输入。然后这些tokens被分别投影为queries，keys，values，denoted as Q，K，V。而注意力图则被定义为A。
% 首先我们调整inpainting区域内容对guidance tokens的关注度，从而使得生成的inpainting内容能够更好的与text prompt在细节上进行对齐。具体来讲，我们将注意力图中Tgf作为key value且inpainting token作为queries的部分进行了调整，希望整个inpainting token在生成过程中能够更多受到guidance tokens的引导，变换后的注意力图如下所示：
%公式
%其中x是图像。
\noindent\textbf{Guided Tokens Enhancement.}
In the stage of the MAR decoder, guidance tokens $T_{gf}$ and all image tokens $I$ are concatenated together as input $X\in \mathbb{R}^{(L+HW)\times D}$. Then these tokens are projected as queries, keys, values, denoted as $Q, K, V\in \mathbb{R}^{(L+HW)\times D'}$, respectively. And the attention map is defined as $A=\frac{QK^T}{\sqrt{D'}}\in \mathbb{R}^{(L+HW)\times (L+HW)}$.
Firstly, we enhance the attention of the inpainting region to guidance tokens, allowing the inpainting content to better align with prompt details. 
% Specifically, we modify the attention scores where the inpainting tokens $I_{p}$ are treated as the queries and the guidance tokens $T_{gf}$ as the keys. 
The enhanced attention map is as follows:
\begin{equation}
\begin{aligned}
A'_{ij}=\begin{cases}\alpha^{\lambda_1}\cdot A_{ij}&X_i\in I_p\ \mathrm{and} \ X_j\in T_{gf},\\
A_{ij}&\mathrm{otherwise},\end{cases}
\end{aligned}
\end{equation}
where $\lambda_1$ is a hyperparameter for the power of $\alpha$.

% 接着,进一步提高区域内容的视觉质量，我们需要使得inpainting区域内的tokens能够更加关注自身信息。具体来讲，我们将注意力图中已预测inpainting token作为keys且未知inpainting token作为queries的部分进行了调整，希望生成过程中未知的inpainting token的内容能够更加受到已预测的inpainting token的引导。不同于guidance tokens和全部inpainting tokens，已预测token与未知token的数量在MAR的生成过程中是动态变化的。因此我们基于MAR的这一生成特性，在每一个生成过程中的step都对增强系数进行调整，具体过程如下所示，
%公式
%其中x是图像。在生成开始时，已预测inpainting token数量较少，因此给予他们较高权重。当已预测token在生成过程中逐渐增多时，权重将会逐渐降低。特别的，在生成开始时，由于没有任何已预测inpainting tokens，因此我们增强整个inpainting区域的注意力图。
\noindent\textbf{Dynamic Inpainting Tokens Enhancement.}
Next, to further enhance the inpainting visual quality, we need to strengthen the interaction within the inpainting tokens. Specifically, we enhance the attention scores of unknown inpainting tokens $I_{p_1}\in \mathbb{R}^{N_1\times D}$ to the predicted inpainting tokens $I_{p_2}\in \mathbb{R}^{N_2\times D}$, where $N_1+N_2=N$. This aims to enable the content of unknown tokens to be more guided by the predicted tokens during the generation process. Unlike guidance tokens, the number of unknown and predicted tokens, \textit{i.e.} $N_1$ and $N_2$, dynamically changes at each step. Therefore, based on this property of MAR, we add an extra coefficient $\beta=\mathrm{log}_{N_2+1}{N_1}$, which dynamically changes at each step.
The enhanced attention map is as follows:
\begin{equation}
\begin{aligned}
A'_{ij}=\begin{cases}\beta^{\lambda_3}\cdot\alpha^{\lambda_2}\cdot A_{ij}&X_i\in I_{p_1}\ \mathrm{and} \ X_j\in I_{p_2},\\A_{ij}&\mathrm{otherwise},\end{cases}
\end{aligned}
\end{equation}
where $\lambda_2$ and $\lambda_3$ is the hyperparameters for the power of $\alpha$ and $\beta$. At the beginning of the generation, the number of predicted inpainting tokens is small, so they are assigned higher weights. As the number of predicted tokens gradually increases, the weights gradually decrease. Specifically, since there are no predicted tokens at the start of generation, we enhance the attention scores of the entire inpainting region.

\section{Experiments}
\subsection{Experimental Settings}
% 为了公平起见，我们选取了最近且最有竞争力的方法。-to-image latent diffusion model的较早工作。Stable Diffusion Inpainting(SDI)是在Stable Diffusion(SD)基础上在随机mask的inpainting数据集上进行微调的专注于text-guided image inpainting任务的模型。HD-Painter和FreeCond则是在SDI基础上进行training-free改进的工作。PowerPaint和BrushNet则是基于SD模型在图像分割得到的inpainting数据集上微调得来。，所有方法都直接或间接的在inpainting数据集上进行过训练。
% add marmeissonic
\textbf{Baseline.} We select recent and competitive methods. SDI is a model fine-tuned on random mask inpainting datasets based on stable diffusion (SD)~\cite{rombach2022high}. HD-Painter~\cite{manukyan2023hd} and FreeCond~\cite{hsiao2024freecond} are training-free improved methods based on SDI. PowerPaint~\cite{powerpaint} and BrushNet~\cite{ju2024brushnet} are fine-tuned on inpainting datasets derived from image segmentation based on SD. Meissonic~\cite{bai2024meissonic} is a MAR model with feature compression layers that employ the naive approach for text-guided inpainting. All diffusion-based methods have been trained on inpainting datasets.

% 为了与这些SOTA方法公平比较，我们从文章中采用了两个最常用text-guided image inpainting任务的数据集。首先，我们在具有宽松mask的EditBench上进行了评测。这个数据集包含240张经过精心标注的图像，每个caption都详细描述inpainting区域内物体的各种属性。与BrushNet不同，我们采用了EditBench中对Mask区域内最为详细的注释，而非整张图片的注释。然后，我们在BrushBench上进行了测评，其包含600个text-image对。与EditBench不同的是，其注释是对整张图片的描述，且mask为类似分割的紧致mask。这两个benchmark可以全面的测试模型在不同类型text prompt和mask下的性能。
\noindent\textbf{Evaluation Benchmarks.} To fairly compare with these methods, we adopt the two most commonly used text-guided image inpainting benchmarks~\cite{ju2024brushnet,wang2023imagen}. First, we evaluate on EditBench, which has loose masks for the inpainting objects. This dataset contains 240 carefully annotated images, with each caption providing a detailed description of the object within the inpainting region. Unlike BrushNet, we use the richest captions of the inpainting regions, rather than the annotations of entire images. Next, we evaluate on BrushBench, which contains 600 text-image pairs. Its captions describe the entire images, and the masks are tight, similar to segmentation masks. These two benchmarks provide a comprehensive evaluation of the performance under different types of prompts and masks.

% 对于Metrics的选择，我们遵循文章从三个方面进行考虑：图像视觉质量，背景区域一致性，文本对齐度。首先，我们采用Image Reward (IR), HPS v2 (HPS), PickScore (PS) and Aesthetic Score (AS)这些与人类偏好对齐的指标。其中 Image Reward (IR), HPS v2 (HPS), PickScore (PS)需要结合text prompt一同计算得分，而Aesthetic Score (AS)仅需单一图像。对于PickScore，我们将每个方法生成的图像和原始图像一同输入，统计生成图像的平均选择率也即得分。然后，对于背景区域一致性，我们选择Peak Signal-to-Noise Ratio (PSNR), Learned Perceptual Image Patch Similarity (LPIPS), and Structural Similarity (SSIM)来测量在生成图片和原始图片间没有mask区域的相似度。最后，我们选择CLIP Similarity (CLIP—S)来在生成的inpainting内容和对应的文本提示间评估文本对齐程度。与过往工作不同，我们首先对生成图像的mask区域的最小外接矩形进行裁剪，然后将这部分与text prompt进行计算来直接评估inpainting区域的文本对齐程度。
\noindent\textbf{Evaluation Metrics.} For the choice of metrics, we follow the previous works~\cite{manukyan2023hd,ju2024brushnet}, considering three aspects: image visual quality, background region consistency, and text alignment. First, we use metrics aligned with human preferences, including Image Reward (IR)~\cite{xu2023imagereward}, HPS v2 (HPS)~\cite{wu2023hpsv}, PickScore (PS)~\cite{kirstain2023pickscore}, and Aesthetic Score (AS)~\cite{schuhmann2022laion_as}. Among these, Image Reward (IR), HPS v2 (HPS), and PickScore (PS) require the text prompt to compute the score, while Aesthetic Score (AS) only needs the image. For PickScore, we input both the generated image and the original image, and calculate the average scores for the generated images. Next, for background region consistency, we select Peak Signal-to-Noise Ratio (PSNR)~\cite{korhonen2012psnr}, Learned Perceptual Image Patch Similarity (LPIPS)~\cite{zhang2018lpips}, and Structural Similarity (SSIM)~\cite{hore2010ssim} to measure the similarity between the generated image and the original image in unmasked regions. Finally, we use CLIP Similarity (CLIP-S)~\cite{wu2021godivaclipscore} to evaluate the text alignment between the generated inpainting content and the text prompt. unlike BrushNet~\cite{ju2024brushnet}, we crop the mask region to compute the scores of alignment with the text prompt.

% 为了确保评估效果的严谨性，我们根据官方释放的代码和模型对所有baselines方法都在两个benchmarks上进行了重新评估，每种方法在所有图像上都使用官方推荐的超参数。为了公平起见，所有的baselines方法都采用参数量为1B的SD-1.5或者SDI-1.5作为base model，而Token Painter则采用NOVA-0.6B作为base model。除非特别说明，本文提及的所有实验都是在NVIDIA L20 GPUs上进行inference。DEIF中，surrounding背景token的比例系数为1，统计量上对齐的系数为0.3，控制调整的高斯函数的形状的系数为250和6。ADAE模块中幂次超参数分别为0.3，0.1和0.03，且该模块仅应用在decoder的前13层来确保前期token交互的增强。
\noindent\textbf{Implementation Details.} To ensure the rigorism of the evaluation, we re-evaluate all baseline methods on the two benchmarks using the officially released code, settings, and models. For fairness, all diffusion-based methods use 0.9B parameter SD-1.5 or SDI-1.5 as base model, and Meissonic use 1B model, while Token Painter uses NOVA-0.6B as base model. All experiments in this paper are conducted on NVIDIA L20 GPUs. In DEIF, the background ratio coefficient $p$ is set to 1, the alignment coefficient $a$ is set to 0.3, and the coefficients controlling the function shape $\varphi$ and $\tau$ are set to 250 and 6. In the ADAE module, the hyperparameters of exponents $\lambda_1$, $\lambda_2$, and $\lambda_3$, are 0.3, 0.1, and 0.03. 
% This module is applied only to the first 13 layers of the decoder to ensure enhanced token interaction in the early stages.

\subsection{Comparisons with State-of-the-Art}
% 表中展示了最近的方法与我们的Token Painter在EditBench和BrushBench上的定量比较。我们的Token Painter展示了非常有竞争力的结果，在两个benchmark的几乎所有指标上都获得了SOTA的结果。 space中进行简单blend的inpainting方法，其简单的拼接做法导致了最差的性能。FreeCond和HD-Painter则是在SDI基础上进行改进的training-free方法，这两种方法在EditBench这种宽松mask上表现出了更高的视觉质量。而在BrushBench这种紧致mask上，这三种方法的视觉质量基本相当，在某些指标上SDI甚至更好一些。PowerPaint和BrushNet这两种直接对stable diffusion在较高质量的inpainting数据集上进行微调的方法则是在所有baseline中最具竞争力的。然而由于diffusion本身的限制，其在局部生成视觉质量和背景一致性上仍不及Token Painter。这是由于MAR在生成过程中，天然具有更好的局部可控性，并且对于背景token不会做任何更改。
% add marmeissonic
\textbf{Quantitative Comparison.}
The Table~\ref{tab:quantitative_comparison} shows that Token Painter demonstrates highly competitive results, achieving state-of-the-art results on almost all metrics across two benchmarks. 
FreeCond and HD-Painter are training-free methods improved upon SDI, showing better visual quality on the loose masks of EditBench, while the visual quality of these three methods is almost the same on the BrushBench with tight masks. PowerPaint and BrushNet, based on SD fine-tuning, perform well across baselines. However, due to the inherent limitations of diffusion, they still lag behind Token Painter in local generation visual quality and background consistency. Meissonic is also competitive, but its naive approach and feature compression layers compromise performance, especially in background consistency.
% The Table~\ref{tab:quantitative_comparison} shows a quantitative comparison between recent methods and our Token Painter on EditBench and BrushBench. Our Token Painter demonstrates highly competitive results, achieving state-of-the-art results on almost all metrics across both benchmarks. uses an early inpainting approach based on simple blending in latent space, and its simplistic stitching approach leads to the worst performance. FreeCond and HD-Painter are training-free methods improved upon SDI, showing better visual quality on the loose masks of EditBench. However, on the tight masks of BrushBench, the visual quality of these three methods is almost the same, with SDI even performing slightly better on some metrics. PowerPaint and BrushNet, which fine-tune Stable Diffusion on higher-quality inpainting datasets, are the most competitive among all the baselines. However, due to the inherent limitations of diffusion models, they still lag behind Token Painter in local generation visual quality and background consistency.

% 与其他inpainting方法的定性比较如图所示，。我们分别从具有宽松mask的EditBench和具有紧致mask的BrushBench中选取了部分样本进行了可视化。Token Painter在色彩，风格，结构和prompt细节对齐方面均表现出了杰出的效果。在第一行中，Token Painter准确跟随了prompt的细节，生成了带有彩色围巾和红色手提箱的羊驼，以及白色帽子和暗黄主体的三头消防栓。而其他方法要么缺失或者混杂了部分元素，要么将生成了与prompt描述完全不一致的物体。在第二行中，Token Painter是唯一一种生成了正确字母形状的方法，展现其优异的局部结构可控性。在第六行中，Token Painter根据周围图像的风格，准确补全了缺失的部分主体，其就如同真实的绘画图像一样。而在其他实例中，Token Painter也在对prompt细节的跟随和局部图像的生成质量上，展现出了与其他基于diffusion方法的明显提升。
% add marmeissonic
\noindent\textbf{Qualitative Comparison.}
The qualitative comparison with other inpainting methods is shown in Figure~\ref{fig:qualitative_comparison}, with the basic SDI excluded due to space limitations. Token Painter demonstrates exceptional performance in color, style, structure, and alignment with prompt details. In the first row, Token Painter accurately follows the prompt details, while Other methods either miss or mix elements. In the second row, Token Painter is the only method that generates the correct letter shapes, showcasing its superior local structural control capability. In the sixth row, Token Painter completes missing parts based on image styles, resembling a real painting. In other examples, Token Painter also shows significant improvements in prompt detail alignment and visual content.
% The qualitative comparison with other inpainting methods is shown in Figure~\ref{fig:qualitative_comparison}, with the worst-performing excluded due to space limitations. Token Painter demonstrates exceptional performance in color, style, structure, and alignment with prompt details. In the first row, Token Painter accurately follows the prompt details, generating a llama with a colorful scarf and a red suitcase, as well as a three-headed fire hydrant with a white cap and dark yellow body. Other methods either miss or mix elements, or generate objects that are completely inconsistent with the prompt description. In the second row, Token Painter is the only method that generates the correct letter shapes, showcasing its superior local structural control. In the sixth row, Token Painter accurately completes the missing parts of the main subject based on the surrounding image style, resembling a real painting. In other examples, Token Painter also shows a clear improvement over other diffusion-based methods in following prompt details and generating high-quality local content.

\begin{figure}[t]
  \centering
   \includegraphics[width=1\linewidth]{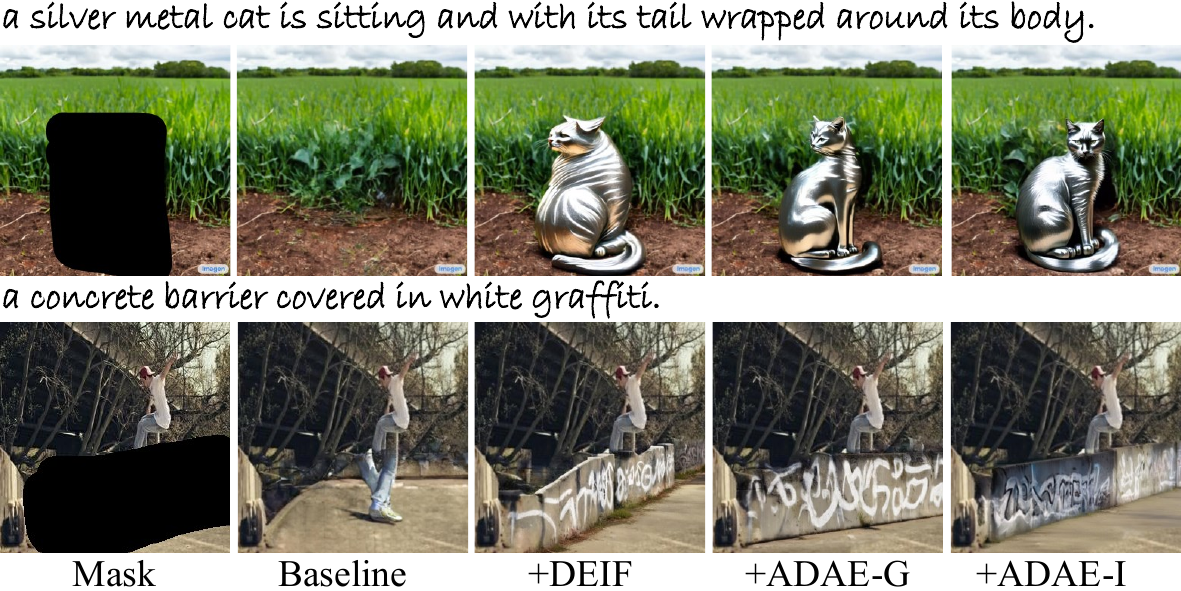}
   \vspace{-1.5em}
   \caption{Visualization of effects of each component. From left to right, we progressively add each proposed component.}
   \vspace{-0.5em}
   \label{fig:ablation_study_vis}
\end{figure}

\tabcolsep=0.12cm
\begin{table}[t]
    \centering
    \begin{tabular}{lcccc}
        \toprule
        Components & IR$_{\times10}$~$\uparrow$ & PS$_{\times10^{2}}$~$\uparrow$ & PSNR~$\uparrow$ & CLIP-S~$\uparrow$ \\
        \midrule
        Baseline  & 4.23 & 19.47 & 26.26 & 6.42 \\
        +DEIF     & 12.41 & 44.26 & 26.35 & 14.42 \\
        +ADAE-G   & 12.76 & 46.28 & 26.27 & 14.45 \\
        +ADAE-I & \textbf{13.01} & \textbf{47.90} & \textbf{26.39} & \textbf{14.46} \\
        \bottomrule
    \end{tabular}
    \vspace{-0.5em}
    \caption{Effects of each component on BrushBench. ADAE-G represents the guidance token enhancement, and ADAE-I is the inpainting token enhancement.}
    \vspace{-1.5em}
    \label{tab:ablation_component}
\end{table}

\begin{figure}[t]
  \centering
   \includegraphics[width=1\linewidth]{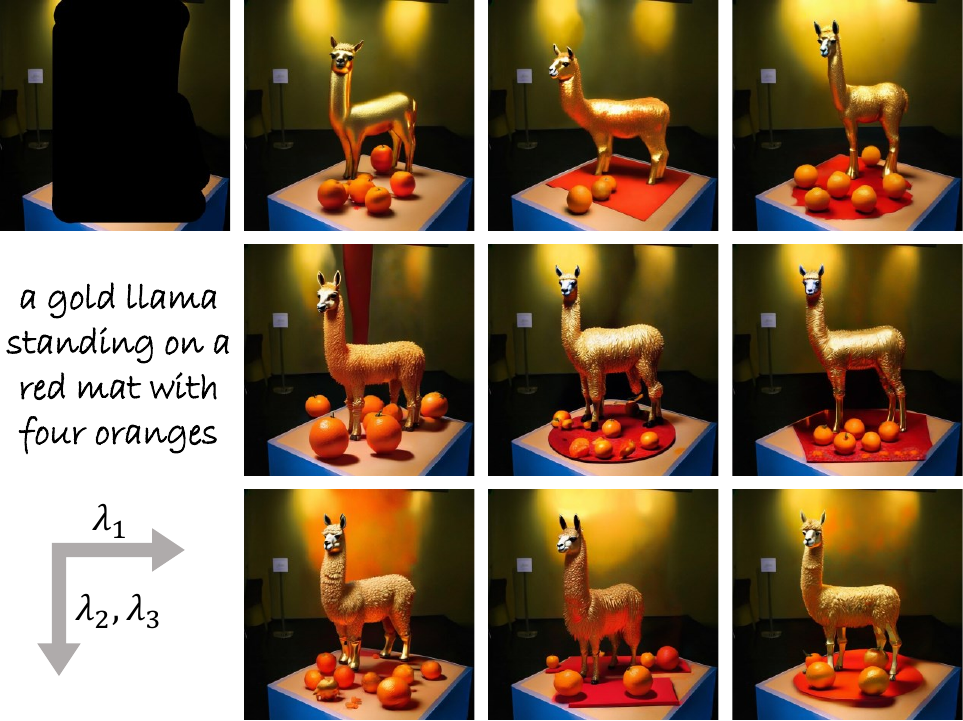}
   \vspace{-1.5em}
   \caption{Effects of two types of power hyperparameters, $\lambda_1$ and $(\lambda_2,\lambda_3)$, varying from 0.1 to 0.3 and (0.03, 0.01) to (0.1, 0.03). The $\lambda_1$ primarily controls the prompt detail alignment, while $(\lambda_2,\lambda_3)$ enhance the content visual quality.}
   \vspace{-0.5em}
   \label{fig:ablation_hyper}
\end{figure}

\tabcolsep=0.12cm
\begin{table}[t]
    \centering
    \begin{tabular}{lcccc}
        \toprule
        Functions & IR$_{\times10}$~$\uparrow$ & PS$_{\times10^{2}}$~$\uparrow$ & PSNR~$\uparrow$ & CLIP-S~$\uparrow$ \\
        \midrule
        Linear    & 12.52 & 44.84 & 26.25 & 14.42 \\
        Constant  & 12.71 & 45.65 & 26.28 & 14.46 \\
        Quadratic & 12.79 & 46.42 & 26.27 & 14.44 \\
        % Gaussian  & 12.92 & 47.09 & 26.35 & 14.43 \\
        M-Gaussian & \textbf{13.01} & \textbf{47.90} & \textbf{26.39} & \textbf{14.46} \\
        \bottomrule
    \end{tabular}
    \vspace{-0.5em}
    \caption{Effects of different functions on BrushBench.}
    \vspace{-1.5em}
    \label{tab:ablation_fusion}
\end{table}

% 为了验证我们方法中每个组件的有效性，我们进行了定量和定性的消融实验。如表和图所示，我们将第一种简单方式作为baseline，即保留background token和text token一同作为输入。我们观察到添加DEIF对模型性能在需要prompt计算的图像质量和CLIP Score上有巨大的提升。这主要是因为DEIF将prompt中的语义信息和背景信息融合得到了一个优质的guidance token，避免了其被背景信息淹没。在添加了ADAE-G之后，inpainting内容对prompt细节的对齐程度进一步提升，这主要得益于增加了inpainting区域对guidance token的关注度。然后我们再添加ADAE-I来使得inpainting区域更关注自身的token信息排除其他无关干扰，使得生成图像质量进行一步提升。我们注意到不同组件下PSNR基本保持不变，这是因为MAR在生成inpainting token不对背景token进行任何更改。
\subsection{Ablation Studies}
\textbf{Effects of Each Component.}
As shown in the Table~\ref{tab:ablation_component} and Figure~\ref{fig:ablation_study_vis}, we use the T\&B approach as the baseline, where we keep both the background token and text token as input. After adding DEIF, we observe a significant improvement in model performance, particularly in image quality and CLIP score. This is mainly because DEIF combines the semantic information from the prompt with the context information from the background to generate novel guidance tokens. After adding ADAE-G, the prompt detail alignment of the inpainting content improves further, primarily due to the increased attention of the inpainting region towards the guidance tokens. We then add ADAE-I to enhance the interaction within inpainting tokens, which leads to a further enhancement in the visual quality. We notice that the PSNR remains almost unchanged across different components, as MAR does not alter background tokens during generation.
% To validate the effectiveness of each component in our method, we conduct both quantitative and qualitative experiments. As shown in the Table~\ref{tab:ablation_component} and Figure~\ref{fig:ablation_study_vis}, we use the T\&B approach as the baseline, where we keep both the background token and text token as input. After adding DEIF, we observe a significant improvement in model performance, particularly in image quality and CLIP score. This is mainly because DEIF combines the semantic information from the prompt with the context information from the background to generate high-quality guidance tokens. After adding ADAE-G, the prompt detail alignment of the inpainting content improves further, primarily due to the increased attention of the inpainting region towards the guidance tokens. We then add ADAE-I to make the inpainting region focus more on its own token information, which leads to a further enhancement in the generated image quality. We notice that the PSNR remains almost unchanged across different components, as MAR does not alter any background tokens during generation.

% 表中显示了不同频域融合函数对性能的影响，这些函数都以L/2为中心，向两边递减。首先我们发现简单的线性函数会带来过多的背景信息，导致图像质量与prompt对齐的下降。然后我们尝试了常数函数，也即在L/4及对称点处进行0-1变换，这种简单的方法虽然提升了prompt的对齐，但是由于粗暴的频域信息拼接，使得图像质量不让人满意。然后我们尝试了二次函数，该函数的形状使得其更多的侧重于语义信息，高频背景信息的缺失限制了图像质量的进一步提升。综上，我们采用了调整后的高斯函数，其在低频段语义信息多高频段背景信息多，同时在L/4处进行平滑过度，使得图像质量和prompt对齐都明显提高。同样的，由于MAR的特性，PSNR基本不变。
\noindent\textbf{Effects of Frequency Fusion Function.}
Table~\ref{tab:ablation_fusion} shows the impact of different frequency fusion functions, all of which are centered at L/2 and decrease symmetrically towards both sides. First, we find that the simple linear function introduces too much context information, leading to a decrease in both image quality and alignment with the prompt. Next, we try a constant function, which performs a 0-1 transformation at L/4. While this simple approach improves prompt alignment, the crude frequency stitching results in unsatisfactory visual quality. We then try a quadratic function, which emphasizes semantic information more, but the lack of high-frequency context information limits further improvement in visual quality. Finally, we adopt a modified Gaussian function, which preserves more semantic information in low-frequency and more context information in high-frequency, while smoothly transitioning at L/4. This results in improvements in both visual quality and prompt alignment. PSNR remains largely unchanged due to the MAR property.

% AEAD模块中主要包括三个超参数，用来控制a和b的幂次大小。这三个超参数可根据其作用分为两种类型。c用来增强inpainting tokens对gudiance token关注度，从而增强生成内容对prompt细节的对齐。d和e则用来增强inpainting tokens对自身的关注度，从而增强物体的视觉质量。如图所示，随着c逐渐加大，inpainting区域在内容上逐渐包含prompt的细节，且结构更加合理。而随着d和e加大，inpainting区域的物体视觉质量逐渐变得更好。然而，实际应用中我们发现过大的幂次超参数会导致inpainting区域变得失真甚至生成混乱的色彩，这可能是由于过度的增强attention分数会导致注意力机制失效。因此我们最终将这些超参数设置为0.3, 0.1和0.03。
\noindent\textbf{Effects of Hyperparameters of Power.}
The ADAE module includes three hyperparameters, $\lambda_1$, $\lambda_2$ and $\lambda_3$, used to control the exponents of $\alpha$ and $\beta$. These hyperparameters are divided into two types. The parameter $\lambda_1$ is used to enhance the attention of inpainting tokens towards the guidance token. The parameters $\lambda_2$ and $\lambda_3$ are used to increase the interaction with inpainting tokens. As shown in the Figure~\ref{fig:ablation_hyper}, the inpainting region gradually incorporates more details from the prompt as $\lambda_1$ increases. When $\lambda_2$ and $\lambda_3$ increase, the visual quality of the objects in the inpainting region improves, and the structure becomes more coherent. However, in practical applications, we find that excessively large power values lead to distortion or even chaotic colors in inpainting regions. This is likely due to the over-enhancement of attention scores, causing the attention mechanism to fail. Therefore, we ultimately set $\lambda_1$, $\lambda_2$, and $\lambda_3$ to 0.3, 0.1, and 0.03.
% The ADAE module includes three hyperparameters, $\lambda_1$, $\lambda_2$ and $\lambda_3$, used to control the exponents of $\alpha$ and $\beta$. These hyperparameters are divided into two types. The parameter $\lambda_3$ is used to enhance the attention of inpainting tokens towards the guidance token, thereby improving the alignment of the inpainting content with the prompt details. The parameters $\lambda_2$ and $\lambda_2$ are used to increase the attention of inpainting tokens towards themselves, improving the visual quality of content. As shown in the Figure~\ref{fig:ablation_hyper}, the inpainting region gradually incorporates more details from the prompt as $\lambda_1$ increases. when $\lambda_2$ and $\lambda_3$ increase, the visual quality of the objects in the inpainting region improves, and the structure becomes more coherent. However, in practical applications, we find that excessively large exponent values lead to distortion in the inpainting region or even generate chaotic colors. This is likely due to the over-enhancement of the attention scores, causing the attention mechanism to fail. Therefore, we ultimately set these hyperparameters to 0.3, 0.1, and 0.03.

\section{Conclusion}
% 在这篇文章中，我们首次将MAR模型应用在text-guided image inpainting任务上，并在深入研究并分析MAR在inpainting任务上的机制后，提出了一种Training-free的框架即Token Painter。它在encoder阶段引入Dual-Stream Encoder Information Fusion (DEIF)模块来获得高质量的inpainting区域的引导特征，在decoder阶段则引入Adaptive Decoder Attention Score Enhancing (ADAE)模块来进一步加强inpainting区域对引导特征和自身区域内token特征的关注度。充分的实验表明，结合这两种在encoder和decoder阶段的改进，Token Painter的性能在几乎所有指标上甚至超过了过往基于微调过的模型（即SDI or SD）的方法，包括SOTA方法。我们希望这篇工作可以促进未来AR模型在image inpainting领域的发展。
In this paper, we improve the T2I MAR model specifically for the text-guided image inpainting task. After conducting analyses of the MAR generation process, we propose a training-free framework, Token Painter. It introduces the Dual-Stream Encoder Information Fusion (DEIF) module at the encoder stage and the Adaptive Decoder Attention Score Enhancing (ADAE) module at the decoder stage. Extensive experiments show that Token Painter outperforms all previous methods, including SOTA methods, across nearly all metrics. We hope that this work will promote the future development of AR models in the image inpainting domain.
% In this paper, we improve the T2I MAR model specifically for the text-guided image inpainting task. After conducting an in-depth study and analysis of MAR’s mechanisms in inpainting, we propose a training-free framework, Token Painter. It introduces the Dual-Stream Encoder Information Fusion (DEIF) module at the encoder stage to obtain high-quality guidance tokens for the inpainting region, and the Adaptive Decoder Attention Score Enhancing (ADAE) module at the decoder stage to further enhance the attention of the inpainting region to both the guidance tokens and the tokens within itself. Extensive experiments show that, by combining these two improvements, Token Painter outperforms previous methods fine-tuned on inpainting datasets, including SOTA methods, across nearly all metrics. We hope that this work will promote the future development of AR models in the image inpainting domain.

\bibliography{aaai2026}

@InProceedings{deJorge_2024_ECCV,
author = {de Jorge, Pau and Volpi, Riccardo and Dakania, Puneet K. and Torr, Philip H. S. and Gregory, Rogez},
title = {Placing Objects in Context via Inpainting for Out-of-distribution Segmentation},
booktitle = {The European Conference on Computer Vision (ECCV)},
month = {October},
year = {2024}
}

@article{li2022misf,
  title={MISF: Multi-level Interactive Siamese Filtering for High-Fidelity Image Inpainting},
  author={Li, Xiaoguang and Guo, Qing and Lin, Di and Li, Ping and Feng, Wei and Wnag, Song},
  journal={CVPR},
  year={2022}
}

@InProceedings{Dong_2022_CVPR,
    author    = {Dong, Qiaole and Cao, Chenjie and Fu, Yanwei},
    title     = {Incremental Transformer Structure Enhanced Image Inpainting With Masking Positional Encoding},
    booktitle = {Proceedings of the IEEE/CVF Conference on Computer Vision and Pattern Recognition (CVPR)},
    month     = {June},
    year      = {2022},
    pages     = {11358-11368}
}

@InProceedings{Liu_2022_CVPR,
    author    = {Liu, Qiankun and Tan, Zhentao and Chen, Dongdong and Chu, Qi and Dai, Xiyang and Chen, Yinpeng and Liu, Mengchen and Yuan, Lu and Yu, Nenghai},
    title     = {Reduce Information Loss in Transformers for Pluralistic Image Inpainting},
    booktitle = {Proceedings of the IEEE/CVF Conference on Computer Vision and Pattern Recognition (CVPR)},
    month     = {June},
    year      = {2022},
    pages     = {11347-11357}
}

@misc{ju2024brushnet,
  title={BrushNet: A Plug-and-Play Image Inpainting Model with Decomposed Dual-Branch Diffusion}, 
  author={Xuan Ju and Xian Liu and Xintao Wang and Yuxuan Bian and Ying Shan and Qiang Xu},
  year={2024},
  eprint={2403.06976},
  archivePrefix={arXiv},
  primaryClass={cs.CV}
}

@misc{powerpaint,
      title={A Task is Worth One Word: Learning with Task Prompts for High-Quality Versatile Image Inpainting},
      author={Junhao Zhuang and Yanhong Zeng and Wenran Liu and Chun Yuan and Kai Chen},
      year={2023},
      eprint={2312.03594},
      archivePrefix={arXiv},
      primaryClass={cs.CV}
}

@article{bld,
  title={Blended latent diffusion},
  author={Avrahami and others},
  journal={ACM transactions on graphics (TOG)},
  volume={42},
  number={4},
  pages={1--11},
  year={2023},
  publisher={ACM New York, NY, USA}
}

@article{nichol2021glide,
  title={Glide: Towards photorealistic image generation and editing with text-guided diffusion models},
  author={Nichol, Alex and Dhariwal, Prafulla and Ramesh, Aditya and Shyam, Pranav and Mishkin, Pamela and McGrew, Bob and Sutskever, Ilya and Chen, Mark},
  journal={arXiv preprint arXiv:2112.10741},
  year={2021}
}

@inproceedings{avrahami2022blended,
  title={Blended diffusion for text-driven editing of natural images},
  author={Avrahami and others},
  booktitle={Proceedings of the IEEE/CVF conference on computer vision and pattern recognition},
  pages={18208--18218},
  year={2022}
}

@article{hsiao2024freecond,
  title={Freecond: Free lunch in the input conditions of text-guided inpainting},
  author={Hsiao, Teng-Fang and Ruan, Bo-Kai and Tsai, Sung-Lin and Wu, Yi-Lun and Shuai, Hong-Han},
  journal={arXiv preprint arXiv:2412.00427},
  year={2024}
}

@inproceedings{manukyan2023hd,
  title={Hd-painter: high-resolution and prompt-faithful text-guided image inpainting with diffusion models},
  author={Manukyan, Hayk and Sargsyan, Andranik and Atanyan, Barsegh and Wang, Zhangyang and Navasardyan, Shant and Shi, Humphrey},
  booktitle={The Thirteenth International Conference on Learning Representations},
  year={2023}
}

@inproceedings{wang2023imagen,
  title={Imagen editor and editbench: Advancing and evaluating text-guided image inpainting},
  author={Wang, Su and Saharia, Chitwan and Montgomery, Ceslee and Pont-Tuset, Jordi and Noy, Shai and Pellegrini, Stefano and Onoe, Yasumasa and Laszlo, Sarah and Fleet, David J and Soricut, Radu and others},
  booktitle={Proceedings of the IEEE/CVF conference on computer vision and pattern recognition},
  pages={18359--18369},
  year={2023}
}

@inproceedings{esser2024scaling,
  title={Scaling rectified flow transformers for high-resolution image synthesis},
  author={Esser, Patrick and Kulal, Sumith and Blattmann, Andreas and Entezari, Rahim and M{\"u}ller, Jonas and Saini, Harry and Levi, Yam and Lorenz, Dominik and Sauer, Axel and Boesel, Frederic and others},
  booktitle={Forty-first international conference on machine learning},
  year={2024}
}

@article{betker2023improving,
  title={Improving image generation with better captions},
  author={Betker, James and Goh, Gabriel and Jing, Li and Brooks, Tim and Wang, Jianfeng and Li, Linjie and Ouyang, Long and Zhuang, Juntang and Lee, Joyce and Guo, Yufei and others},
  journal={Computer Science. https://cdn. openai. com/papers/dall-e-3. pdf},
  pages={8},
  year={2023}
}

@article{saharia2022photorealistic,
  title={Photorealistic text-to-image diffusion models with deep language understanding},
  author={Saharia, Chitwan and Chan, William and Saxena, Saurabh and Li, Lala and Whang, Jay and Denton, Emily L and Ghasemipour, Kamyar and Gontijo Lopes, Raphael and Karagol Ayan, Burcu and Salimans, Tim and others},
  journal={Advances in neural information processing systems},
  volume={35},
  pages={36479--36494},
  year={2022}
}

@inproceedings{rombach2022high,
  title={High-resolution image synthesis with latent diffusion models},
  author={Rombach, Robin and Blattmann, Andreas and Lorenz, Dominik and Esser, Patrick and Ommer, Bj{\"o}rn},
  booktitle={Proceedings of the IEEE/CVF conference on computer vision and pattern recognition},
  pages={10684--10695},
  year={2022}
}

@article{chen2023pixart,
  title={Pixart-$alpha$: Fast training of diffusion transformer for photorealistic text-to-image synthesis},
  author={Chen, Junsong and Yu, Jincheng and Ge, Chongjian and Yao, Lewei and Xie, Enze and Wu, Yue and Wang, Zhongdao and Kwok, James and Luo, Ping and Lu, Huchuan and others},
  journal={arXiv preprint arXiv:2310.00426},
  year={2023}
}

@article{sun2024autoregressive,
  title={Autoregressive model beats diffusion: Llama for scalable image generation},
  author={Sun, Peize and Jiang, Yi and Chen, Shoufa and Zhang, Shilong and Peng, Bingyue and Luo, Ping and Yuan, Zehuan},
  journal={arXiv preprint arXiv:2406.06525},
  year={2024}
}

@article{team2024chameleon,
  title={Chameleon: Mixed-modal early-fusion foundation models},
  author={Team, Chameleon},
  journal={arXiv preprint arXiv:2405.09818},
  year={2024}
}

@article{yu2022scaling,
  title={Scaling autoregressive models for content-rich text-to-image generation},
  author={Yu, Jiahui and Xu, Yuanzhong and Koh, Jing Yu and Luong, Thang and Baid, Gunjan and Wang, Zirui and Vasudevan, Vijay and Ku, Alexander and Yang, Yinfei and Ayan, Burcu Karagol and others},
  journal={arXiv preprint arXiv:2206.10789},
  volume={2},
  number={3},
  pages={5},
  year={2022}
}

@article{tian2024visual,
  title={Visual autoregressive modeling: Scalable image generation via next-scale prediction},
  author={Tian, Keyu and Jiang, Yi and Yuan, Zehuan and Peng, Bingyue and Wang, Liwei},
  journal={Advances in neural information processing systems},
  volume={37},
  pages={84839--84865},
  year={2024}
}

@article{fan2410fluid,
  title={Fluid: Scaling autoregressive text-to-image generative models with continuous tokens, 2024},
  author={Fan, Lijie and Li, Tianhong and Qin, Siyang and Li, Yuanzhen and Sun, Chen and Rubinstein, Michael and Sun, Deqing and He, Kaiming and Tian, Yonglong},
  journal={URL https://arxiv. org/abs/2410.13863},
  year={2024}
}

@inproceedings{pang2025randar,
  title={Randar: Decoder-only autoregressive visual generation in random orders},
  author={Pang, Ziqi and Zhang, Tianyuan and Luan, Fujun and Man, Yunze and Tan, Hao and Zhang, Kai and Freeman, William T and Wang, Yu-Xiong},
  booktitle={Proceedings of the Computer Vision and Pattern Recognition Conference},
  pages={45--55},
  year={2025}
}

@article{yu2025frequency,
  title={Frequency autoregressive image generation with continuous tokens},
  author={Yu, Hu and Luo, Hao and Yuan, Hangjie and Rong, Yu and Zhao, Feng},
  journal={arXiv preprint arXiv:2503.05305},
  year={2025}
}

@article{ho2020denoising,
  title={Denoising diffusion probabilistic models},
  author={Ho, Jonathan and Jain, Ajay},
  journal={Advances in neural information processing systems},
  volume={33},
  pages={6840--6851},
  year={2020}
}

@article{song2020score,
  title={Score-based generative modeling through stochastic differential equations},
  author={Song, Yang and Sohl-Dickstein, Jascha and Kingma, Diederik P and Kumar, Abhishek and Ermon, Stefano and Poole, Ben},
  journal={arXiv preprint arXiv:2011.13456},
  year={2020}
}

@article{vaswani2017attention,
  title={Attention is all you need},
  author={Vaswani, Ashish and Shazeer, Noam and Parmar, Niki and Uszkoreit, Jakob and Jones, Llion and Gomez, Aidan N and Kaiser, {\L}ukasz and Polosukhin, Illia},
  journal={Advances in neural information processing systems},
  volume={30},
  year={2017}
}

@article{radford2018improving,
  title={Improving Language Understanding by Generative Pre-Training},
  author={Radford, Alec and Narasimhan, Karthik and Salimans, Tim and Sutskever, Ilya and others},
  year={2018},
  journal={https://api.semanticscholar.org/CorpusID:49313245},
  publisher={San Francisco, CA, USA}
}

@inproceedings{chang2022maskgit,
  title={Maskgit: Masked generative image transformer},
  author={Chang, Huiwen and Zhang, Han and Jiang, Lu and Liu, Ce and Freeman, William T},
  booktitle={Proceedings of the IEEE/CVF conference on computer vision and pattern recognition},
  pages={11315--11325},
  year={2022}
}

@article{li2024autoregressive,
  title={Autoregressive image generation without vector quantization},
  author={Li, Tianhong and Tian, Yonglong and Li, He and Deng, Mingyang and He, Kaiming},
  journal={Advances in Neural Information Processing Systems},
  volume={37},
  pages={56424--56445},
  year={2024}
}

@inproceedings{bai2024meissonic,
  title={Meissonic: Revitalizing masked generative transformers for efficient high-resolution text-to-image synthesis},
  author={Bai, Jinbin and Ye, Tian and Chow, Wei and Song, Enxin and Chen, Qing-Guo and Li, Xiangtai and Dong, Zhen and Zhu, Lei and Yan, Shuicheng},
  booktitle={The Thirteenth International Conference on Learning Representations},
  year={2024}
}

@article{chang2023muse,
  title={Muse: Text-to-image generation via masked generative transformers},
  author={Chang, Huiwen and Zhang, Han and Barber, Jarred and Maschinot, AJ and Lezama, Jose and Jiang, Lu and Yang, Ming-Hsuan and Murphy, Kevin and Freeman, William T and Rubinstein, Michael and others},
  journal={arXiv preprint arXiv:2301.00704},
  year={2023}
}

@article{deng2024autoregressive,
  title={Autoregressive video generation without vector quantization},
  author={Deng, Haoge and Pan, Ting and Diao, Haiwen and Luo, Zhengxiong and Cui, Yufeng and Lu, Huchuan and Shan, Shiguang and Qi, Yonggang and Wang, Xinlong},
  journal={arXiv preprint arXiv:2412.14169},
  year={2024}
}

@article{jin2023training,
  title={Training-free diffusion model adaptation for variable-sized text-to-image synthesis},
  author={Jin, Zhiyu and Shen, Xuli and Li, Bin and Xue, Xiangyang},
  journal={Advances in Neural Information Processing Systems},
  volume={36},
  pages={70847--70860},
  year={2023}
}

@article{sun2024uniavatar,
  title={Uniavatar: Taming lifelike audio-driven talking head generation with comprehensive motion and lighting control},
  author={Sun, Wenzhang and Li, Xiang and Di, Donglin and Liang, Zhuding and Zhang, Qiyuan and Li, Hao and Chen, Wei and Cui, Jianxun},
  journal={arXiv preprint arXiv:2412.19860},
  year={2025}
}

@inproceedings{liu2025moee,
  title={Moee: Mixture of emotion experts for audio-driven portrait animation},
  author={Liu, Huaize and Sun, Wenzhang and Di, Donglin and Sun, Shibo and Yang, Jiahui and Zou, Changqing and Bao, Hujun},
  booktitle={Proceedings of the Computer Vision and Pattern Recognition Conference},
  pages={26222--26231},
  year={2025}
}

@article{xu2023imagereward,
  title={Imagereward: Learning and evaluating human preferences for text-to-image generation},
  author={Xu, Jiazheng and Liu, Xiao and Wu, Yuchen and Tong, Yuxuan and Li, Qinkai and Ding, Ming and Tang, Jie and Dong, Yuxiao},
  journal={Advances in Neural Information Processing Systems},
  volume={36},
  pages={15903--15935},
  year={2023}
}

@article{wu2023hpsv,
  title={Human preference score v2: A solid benchmark for evaluating human preferences of text-to-image synthesis},
  author={Wu, Xiaoshi and Hao, Yiming and Sun, Keqiang and Chen, Yixiong and Zhu, Feng and Zhao, Rui and Li, Hongsheng},
  journal={arXiv preprint arXiv:2306.09341},
  year={2023}
}

@article{kirstain2023pickscore,
  title={Pick-a-pic: An open dataset of user preferences for text-to-image generation},
  author={Kirstain, Yuval and Polyak, Adam and Singer, Uriel and Matiana, Shahbuland and Penna, Joe and Levy, Omer},
  journal={Advances in neural information processing systems},
  volume={36},
  pages={36652--36663},
  year={2023}
}

@article{schuhmann2022laion_as,
  title={Laion-5b: An open large-scale dataset for training next generation image-text models},
  author={Schuhmann, Christoph and Beaumont, Romain and Vencu, Richard and Gordon, Cade and Wightman, Ross and Cherti, Mehdi and Coombes, Theo and Katta, Aarush and Mullis, Clayton and Wortsman, Mitchell and others},
  journal={Advances in neural information processing systems},
  volume={35},
  pages={25278--25294},
  year={2022}
}

@inproceedings{korhonen2012psnr,
  title={Peak signal-to-noise ratio revisited: Is simple beautiful?},
  author={Korhonen, Jari and You, Junyong},
  booktitle={2012 Fourth international workshop on quality of multimedia experience},
  pages={37--38},
  year={2012},
  organization={IEEE}
}

@inproceedings{zhang2018lpips,
  title={The unreasonable effectiveness of deep features as a perceptual metric},
  author={Zhang, Richard and Isola, Phillip and Efros, Alexei A and Shechtman, Eli and Wang, Oliver},
  booktitle={Proceedings of the IEEE conference on computer vision and pattern recognition},
  pages={586--595},
  year={2018}
}

@inproceedings{hore2010ssim,
  title={Image quality metrics: PSNR vs. SSIM},
  author={Hore, Alain and Ziou, Djemel},
  booktitle={2010 20th international conference on pattern recognition},
  pages={2366--2369},
  year={2010},
  organization={IEEE}
}

@article{wu2021godivaclipscore,
  title={Godiva: Generating open-domain videos from natural descriptions},
  author={Wu, Chenfei and Huang, Lun and Zhang, Qianxi and Li, Binyang and Ji, Lei and Yang, Fan and Sapiro, Guillermo and Duan, Nan},
  journal={arXiv preprint arXiv:2104.14806},
  year={2021}
}

@inproceedings{kwon2024aesfa,
  title={Aesfa: an aesthetic feature-aware arbitrary neural style transfer},
  author={Kwon, Joonwoo and Kim, Sooyoung and Lin, Yuewei and Yoo, Shinjae and Cha, Jiook},
  booktitle={Proceedings of the AAAI conference on artificial intelligence},
  volume={38},
  pages={13310--13319},
  year={2024}
}

@inproceedings{gao2024frequency,
  title={Frequency-controlled diffusion model for versatile text-guided image-to-image translation},
  author={Gao, Xiang and Xu, Zhengbo and Zhao, Junhan and Liu, Jiaying},
  booktitle={Proceedings of the AAAI Conference on Artificial Intelligence},
  volume={38},
  pages={1824--1832},
  year={2024}
}
\clearpage


\section*{Supplementary material}

# AAAI 2026 Supplementary Material
# Camera Ready

**Longtao Jiang[1], Jie Huang[1], Mingfei Han[2], Lei Chen[1],**
**Yongqiang Yu[2], Feng Zhao[1], Xiaojun Chang[1], Zhihui Li[1†]**

[1]University of Science and Technology of China  [2]Department of Computer Vision, MBZUAI
{taotao707,hj0117}@mail.ustc.edu.cn,  mingfei.han,yongqiang.yu@mbzuai.ac.ae
chenlei.hfut@gmail,  {fzhao956,xjchang,lizhihuics}@ustc.edu.cn

## 1  More Baselines Comparison

Due to space limitations, we omit two classic text-guided inpainting methods, Blended latent diffusion(BLD) (Avrahami et al. 2023) and ControlNet-Inpainting(CNI) (Zhang, Rao, and Agrawala 2023) in the main paper. Here, we provide their quantitative evaluation results as a supplement, as shown in Table 1. BLD is an early work that performs inpainting by simple blending in the latent space, its straightforward approach results in poor performance. CNI applies the ControlNet branch, originally designed to control fine spatial structures in diffusion models, to the inpainting task, injecting the masked image through ControlNet to preserve the original background. Since it is also based on Stable Diffusion 1.5 (Rombach et al. 2022), like Stable Diffusion Inpainting (SDI), it achieves performance comparable to SDI in terms of image quality and text alignment. However, its coarse background injection leads to worse background consistency compared to SDI. Compared with the above two methods, our Token Painter outperforms both methods across all metrics on both benchmarks.

It is worth noting that both Muse (Chang et al. 2023) and Meissonic (Bai et al. 2024) involve in the text-guided image inpainting task by native approaches that simply empty the inpainting tokens. However, since Muse is a relatively early work and, more importantly, not open-sourced, we report only the results of Meissonic in the main paper. Given the architectural similarity between Muse and Meissonic, and the fact that Meissonic demonstrates competitive performance compared to Muse in its paper, we believe that our Token Painter is also likely to be more competitive than Muse.

## 2  Attention Maps after Token Painter

To further illustrate the impact of our Token Painter on the generation process, we visualize two types of attention maps after incorporating the two key components: the DEIF module and the ADAE module, as shown in the Figure 1. After adding the DEIF module, we observe that, compared to the simpler T&B and T-only approaches, most of the attention is concentrated around the inpainting region, with relatively little attention on the background. This is because the DEIF module fuses semantic and contextual information, enabling

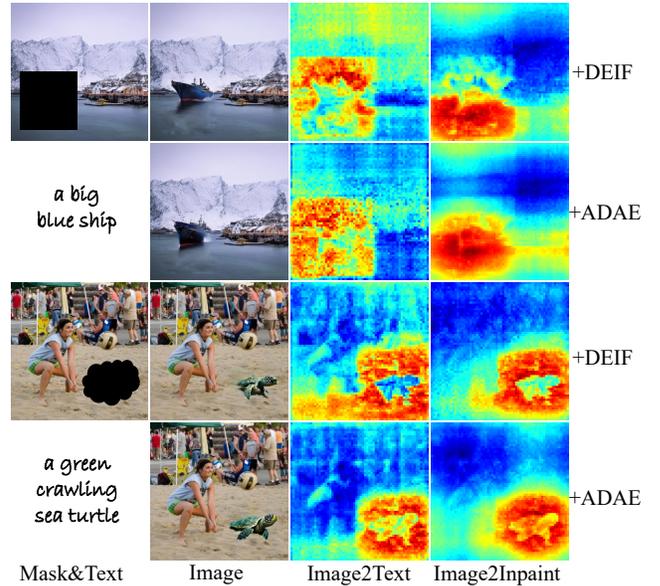

Figure 1: Visualization of two types of attention maps after Token Painter. DEIF indicates that after adding the DEIF module to the baseline, ADAE indicates that the ADAE module has been further added.

the inpainting content to both follow the text prompt and seamlessly connect with the surrounding context. As a result, the attention is reasonably distributed across the inpainting region and the nearby contextual regions.

After adding the ADAE module, we observe that while a small amount of attention remains on the background, the attention further concentrates on the inpainting region, particularly on the object region generated according to the text prompt. This indicates that with the ADAE module, the model focuses more on generating the object content within the inpainting region, further enhancing both visual quality and alignment with prompt details.

To visualize the attention maps, we extract the self-attention maps from each layer of the decoder at step $K$, when all tokens have been fully generated. Based on the positions of the guidance tokens, $i.e.$, the updated text tokens, and the inpainting tokens, we extract two corresponding attention maps: using image tokens as queries and the

| Dataset | Methods | IR$_{\times 10}$ ↑ | PS$_{\times 10^2}$ ↑ | HPS$_{\times 10^2}$ ↑ | AS ↑ | PSNR ↑ | LPIPS$_{\times 10^3}$ ↓ | SSIM$_{\times 10}$ ↑ | CLIP-S ↑ |
|---|---|---|---|---|---|---|---|---|---|
| EditBench | BLD | -9.78 | 38.23 | 19.68 | 3.57 | 18.52 | 79.53 | 7.09 | 24.00 |
| | CNI | -9.72 | 39.42 | 21.28 | 3.85 | 23.76 | 26.19 | 8.39 | 23.61 |
| | Token Painter (Ours) | **-2.49** | **55.37** | **23.00** | **3.92** | **28.03** | **24.92** | **9.41** | **26.06** |
| BrushBench | BLD | 8.60 | 30.76 | 24.29 | 4.11 | 19.58 | 133.50 | 6.67 | 14.41 |
| | CNI | 11.56 | 39.34 | 27.29 | 4.19 | 22.68 | 44.40 | 7.66 | 14.43 |
| | Token Painter (Ours) | **13.01** | **47.90** | **28.36** | **4.22** | **26.39** | **42.27** | **8.78** | **14.46** |

Table 1: Quantitative results on the EditBench and BrushBench datasets. The **best results** are marked in bold and underline, respectively. Results are all re-evaluated based on the released code, models and setting.

| Methods | Backbone | Params(B) | Speed(s) |
|---|---|---|---|
| BLD | LDM | 0.672 | 4.08 |
| SDI | SDI-1.5 | 0.860 | 6.67 |
| CNI | SD-1.5 | 1.221 | 9.20 |
| HD-Painter | SDI-1.5 | 0.860 | 13.16 |
| PowerPaint | SD-1.5 | 0.860 | 6.85 |
| BrushNet | SD-1.5 | 1.478 | 14.27 |
| FreeCond | SDI-1.5 | 0.860 | 6.89 |
| Meissonic | Meissonic-1B | 1.007 | 13.45 |
| Token Painter | NOVA-0.6B | 0.646 | 12.77 |

Table 2: Comparison of the efficiency of different methods with Token Painter. We calculated the number of model parameters and the average inference time for a single image.

two types of tokens as keys. We then average along the key dimension and resize the result to $H \times W$ for visualization.

## 3 Computational Efficiency Analysis

To better demonstrate the efficiency of our method Token Painter compared to others, we report the number of model parameters and the average inference time per image on BrushBench for each method, as shown in the Table 2. All experiments are conducted on NVIDIA L20 GPUs. For all diffusion-based methods, we set the number of denoising steps to the standard 50. For Meissonic and our method Token Painter, we set the number of generation steps to the 32, which is the half of the text-to-image task.

We observe that BLD, based on the early LDM(Rombach et al. 2022), has the lowest spatial and temporal complexity, but this also results in the worst performance. FreeCond, which introduces simple training-free improvements to SDI, maintains roughly the same time complexity. Although HD-Painter is also a training-free improvement over SDI, its RASG module modifies the sampling strategy and introduces additional gradient computations, leading to increased time complexity. PowerPaint is retrained based on SD-1.5 and introduces learnable semantic tokens, since its inference process is largely consistent with standard SDI, it achieves similar time complexity. Additionally, since no extra branch network structures are introduced, these methods share the same spatial complexity.

CNI and BrushNet, though also based on SD-1.5, introduce additional network branches, which lead to varying degrees of increased time complexity and spatial complexity.

For Meissonic and our method, Token Painter, both follow the text-to-image generation pipeline with 64 steps for generating the full image. However, since we only need to generate part of the image, we set the step count to 32, consistent with the setup in the main paper. While Token Painter involves operations such as frequency-domain fusion, its underlying NOVA backbone has fewer parameters, resulting in slightly lower time complexity than Meissonic.

In summary, compared to prior methods, Token Painter achieves lower spatial complexity and maintains competitive time complexity. Given its superior visual generation quality and alignment with text prompts, our method stands out as a strong text-guided image inpainting solution.

## 4 More Qualitative Results

Due to space limitations in the main paper, we omit the qualitative results of SDI and also reduce the size of other visualization figures. Here, we provide additional and higher-resolution qualitative visualizations, including baseline methods such as SDI, CNI, and BLD. These images are chosen from both EditBench and BrushBench, as shown in the Figure 2 and Figure 3. These visualizations demonstrate the outstanding performance of Token Painter.

## References

Avrahami; et al. 2023. Blended latent diffusion. *ACM transactions on graphics (TOG)*, 42(4): 1–11.

Bai, J.; Ye, T.; Chow, W.; Song, E.; Chen, Q.-G.; Li, X.; Dong, Z.; Zhu, L.; and Yan, S. 2024. Meissonic: Revitalizing masked generative transformers for efficient high-resolution text-to-image synthesis. In *The Thirteenth International Conference on Learning Representations*.

Chang, H.; Zhang, H.; Barber, J.; Maschinot, A.; Lezama, J.; Jiang, L.; Yang, M.-H.; Murphy, K.; Freeman, W. T.; Rubinstein, M.; et al. 2023. Muse: Text-to-image generation via masked generative transformers. *arXiv preprint arXiv:2301.00704*.

Rombach, R.; Blattmann, A.; Lorenz, D.; Esser, P.; and Ommer, B. 2022. High-resolution image synthesis with latent diffusion models. In *Proceedings of the IEEE/CVF conference on computer vision and pattern recognition*, 10684–10695.

Zhang, L.; Rao, A.; and Agrawala, M. 2023. Adding conditional control to text-to-image diffusion models. In *Proceedings of the IEEE/CVF international conference on computer vision*, 3836–3847.

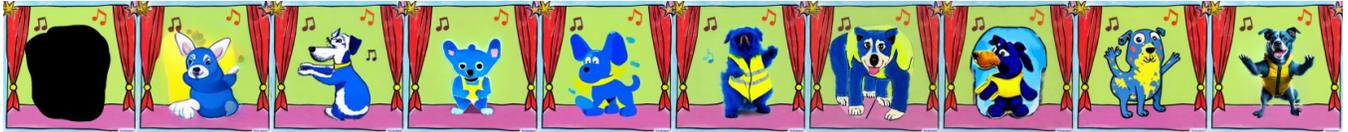
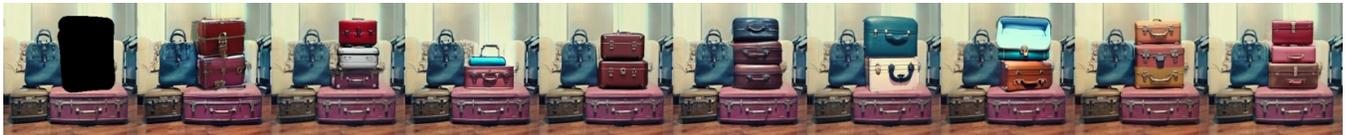
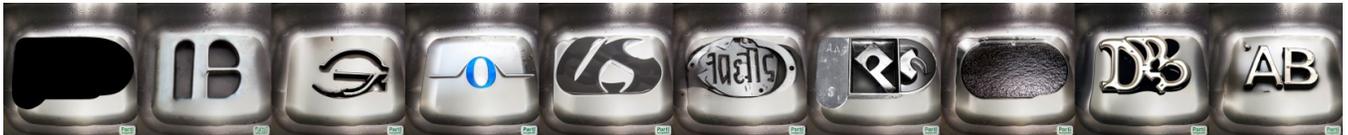
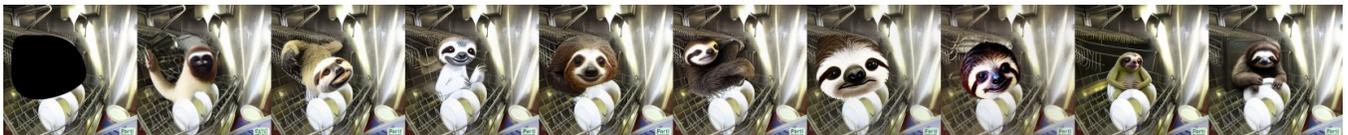
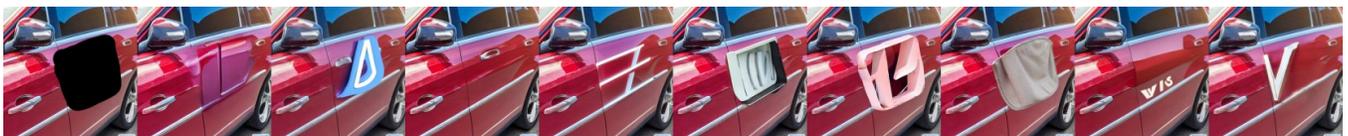
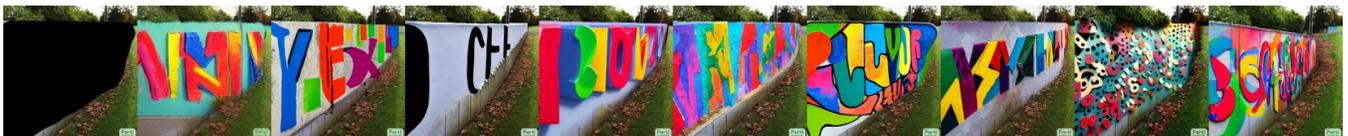
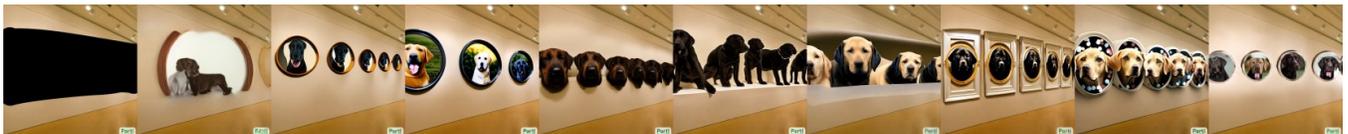
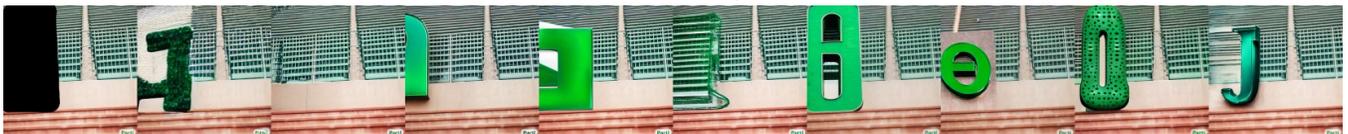
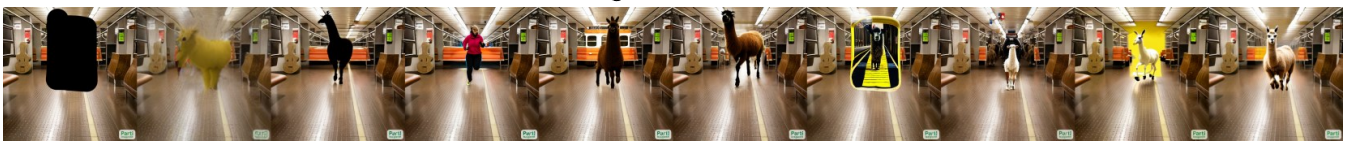

Figure 2: Qualitative results of our Token Painter with previous text-guided inpainting methods on EditBench.

a plate of dumplings with chopsticks on top

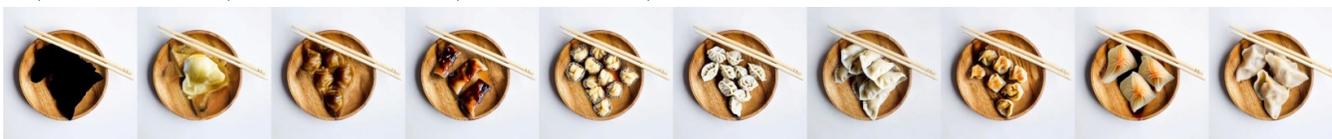

a house in the middle of a field at sunset

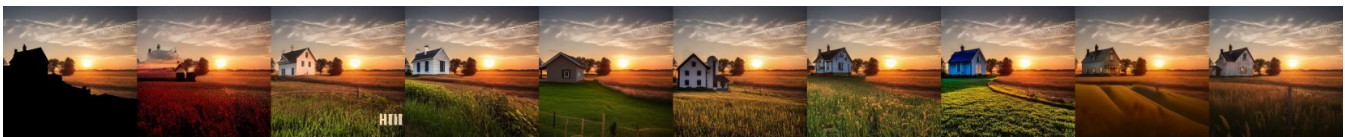

a painting of a cat with stars and a golden orb

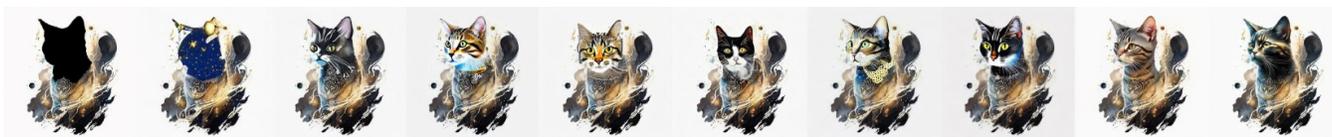

a cat sitting on top of a blue post

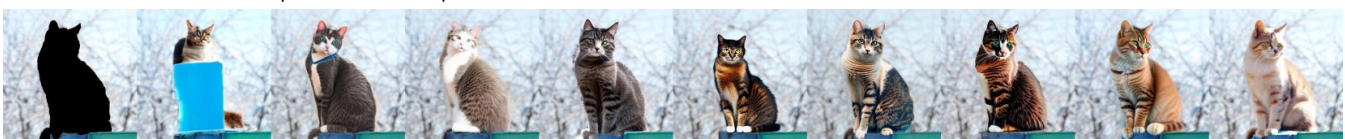

a woman in a white dress holding a fan

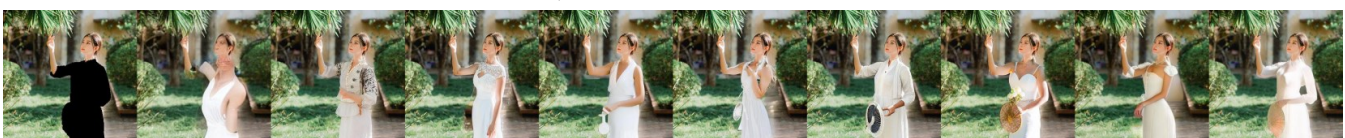

two tents in the desert with a car parked in the background

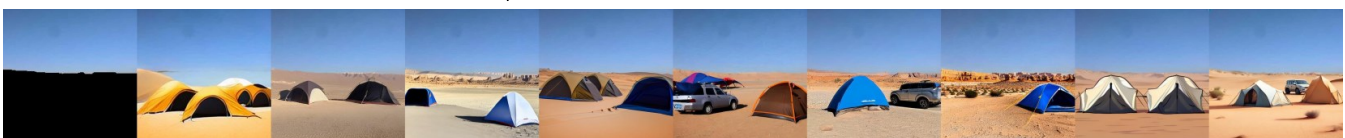

a ship in a bottle on the ocean

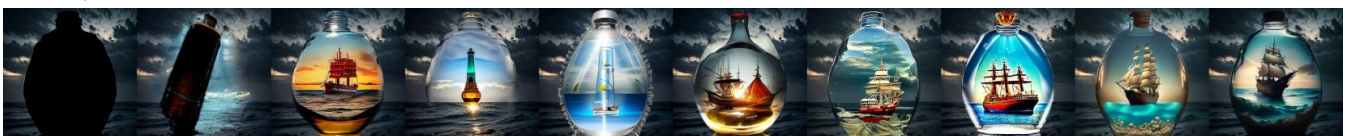

a red barn in the snow with trees

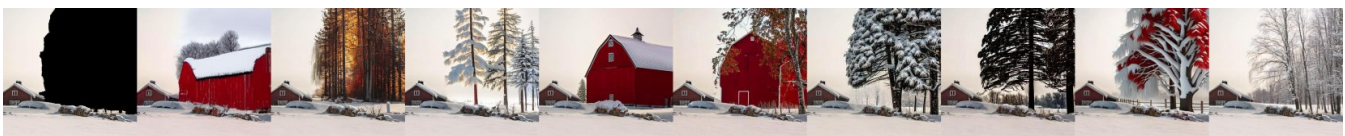

a painting of planets in the sky with stars

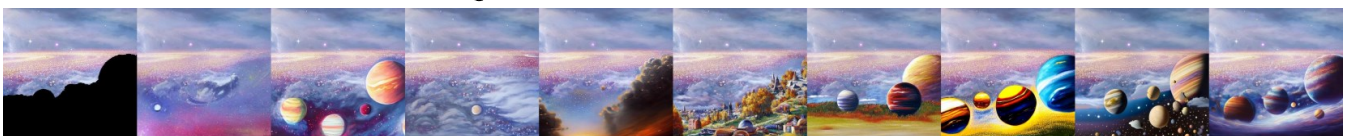

    Mask    BLD    SDI    CNI    HD-Painter  PowerPaint  BrushNet  FreeCond  Meissonic  Ours

Figure 3: Qualitative results of our Token Painter with previous text-guided inpainting methods on BrushBench.


\end{document}